\documentclass{article}
\usepackage{amsmath}
\usepackage{psfrag}
\usepackage{hyperref}
\usepackage{import}
\usepackage{subcaption}
\usepackage{ragged2e}
\usepackage{comment}
\usepackage{algorithm}
\usepackage[noend]{algpseudocode}

\usepackage{xcolor}			
\usepackage{color}			
\usepackage{transparent}

\usepackage[normalem]{ulem}

\usepackage{setspace}

\newcommand{\bs}{\boldsymbol}

\newcommand{\refeq}[1]{Equation \eqref{#1}}

\newcommand{\ee}{\end{equation}}
\newcommand{\be}{\begin{equation}}
\newcommand{\ec}{\end{center}}
\newcommand{\bc}{\begin{center}}
\newcommand{\eea}{\end{eqnarray}}
\newcommand{\bea}{\begin{eqnarray}}
\newcommand{\bd}{\begin{description}}
\newcommand{\ed}{\end{description}}
\newcommand{\bi}{\begin{itemize}}
\newcommand{\ei}{\end{itemize}}

\newcommand\bksi{\boldsymbol{\xi}}

\newcommand{\bx}{\bs{x}}

\newcommand{\by}{\bs{y}}

\newcommand{\bz}{\bs{z}}

\newcommand{\bxx}{\bs{X}}

\newcommand{\byy}{\bs{Y}}

\newcommand{\bt}{\bs{\theta}}

\usepackage{cancel}
\usepackage{ulem}
\usepackage{booktabs}
\usepackage{amssymb}
\usepackage{PRIMEarxiv}

\usepackage[utf8]{inputenc} 
\usepackage[T1]{fontenc}    
\usepackage{hyperref}       
\usepackage{url}            
\usepackage{booktabs}       
\usepackage{amsfonts}       
\usepackage{nicefrac}       
\usepackage{microtype}      
\usepackage{lipsum}
\usepackage{graphicx}
\graphicspath{{media/}}     

\begin{document}

\title{
GenPANIS: A Latent-Variable Generative Framework for Forward and Inverse PDE Problems in Multiphase Media \thanks{\textit{Corresponding author\\
Email addresses: matthaios.chatzopoulos@tum.de (Matthaios Chatzopoulos), \\
p.s.koutsourelakis@tum.de (Phaedon-Stelios Koutsourelakis)}}}

\author{
  Matthaios Chatzopoulos$^a$, Phaedon-Stelios Koutsourelakis$^{a, b, *}$ \\
  $^a$ Technical University of Munich, Professorship of Data-driven Materials Modeling,\\ School of
Engineering and Design, Boltzmannstr. 15, 85748 Garching, Germany \\
$^b$ Munich Data Science Institute (MDSI - Core member), Garching, Germany}

\maketitle

\begin{abstract}

Inverse problems and inverse design in multiphase media, i.e., recovering or engineering microstructures to achieve target macroscopic responses, require operating on discrete-valued material fields, rendering the problem non-differentiable and incompatible with gradient-based methods. Existing approaches either relax to continuous approximations, compromising physical fidelity, or employ separate heavyweight models for forward and inverse tasks. We propose GenPANIS, a unified generative framework that preserves exact discrete microstructures while enabling gradient-based inference through continuous latent embeddings. The model learns a joint distribution over microstructures and PDE solutions, supporting bidirectional inference (forward prediction and inverse recovery) within a single architecture.
The generative formulation enables training with unlabeled data, physics residuals, and minimal labeled pairs. A physics-aware decoder incorporating a differentiable coarse-grained PDE solver preserves governing equation structure, enabling extrapolation to varying boundary conditions and microstructural statistics.
A learnable normalizing flow prior captures complex posterior structure for inverse problems. 
Demonstrated on Darcy flow and Helmholtz equations, GenPANIS maintains accuracy on challenging extrapolative scenarios—including unseen boundary conditions, volume fractions, and microstructural morphologies, with sparse, noisy observations. It outperforms state-of-the-art methods while using $10–100$ times fewer parameters and providing principled uncertainty quantification.

\end{abstract}


\section{Introduction}
\label{introduction}
Heterogeneous media are ubiquitous in science and engineering, arising in porous and composite materials, multiphase solids, and structured thermal, acoustic, or electromagnetic systems \cite{torquato2002random}. In such media, material properties vary sharply in space and are often represented as discrete-valued fields corresponding to distinct phases or constituents. The microscale arrangement of these phases plays a decisive role in determining macroscopic behavior \cite{yvonnet2019computational, keller2011effective}, giving rise to intrinsically high-dimensional and multiscale problems. Beyond prediction, many applications require inverse analysis and inverse design: inferring or engineering microstructures that produce desired macroscopic responses under prescribed conditions.

Inverse problems in multiphase media have  been approached using deterministic optimization \cite{HegemannCantareroRichardsonTeran2013} or Bayesian formulations \cite{kaipio2005computational}. Both paradigms encounter a fundamental obstruction stemming from the discrete nature of material fields: when coefficients take values from a finite set, the forward map becomes non-differentiable with respect to the microstructure. This precludes gradient-based methods, forcing reliance on  relaxations, combinatorial searches, or derivative-free sampling schemes that scale poorly with dimensionality \cite{iglesias2014well}.

To mitigate computational cost, surrogate models have been widely adopted \cite{piao2024domain}. Most existing approaches treat forward and inverse problems asymmetrically: forward surrogates approximate the solution operator, while inverse problems are addressed separately through optimization, sampling, or learned inverse mappings \cite{li2021physics, li2020fourier, stankevich2021numerical, anantha_padmanabha_solving_2021}. This separation is problematic because inverse problems fundamentally differ in structure: the forward operator smooths input fields, whereas inverse inference must recover sharp, discrete structures from partial and noisy observations \cite{isakov2006inverse, chung2017generalized, xu2025numerical}.

A common workaround is to replace discrete material fields with continuous relaxations \cite{ciarbonetti_identification_2022, li2021physics, LinHe2025Differentiable}. While enabling differentiation, these methods fundamentally alter the problem being solved: inferred solutions are no longer discrete-valued and require post hoc thresholding or projection to recover admissible configurations, compromising physical fidelity and uncertainty quantification.
Alternative approaches employ adaptive surrogate schemes \cite{CuiMW16, turn0search0, yan2018adaptivePC, turn0search5} that iteratively refine models based on posterior information but require repeated PDE evaluations. Self-adaptive PINNs \cite{aziz2025self} embed unknown coefficients as trainable parameters, preserving interfaces but lacking principled uncertainty quantification. Direct inverse mapping approaches \cite{molinaro2023neural, kaltenbach2023semi, stankevich2021numerical} learn fast conditional maps but cannot handle varying sensor configurations, noise levels, or boundary conditions at test time.

Generative models offer a more flexible paradigm. Variational autoencoders have been applied to reduce dimensionality for Bayesian inversion \cite{laloy2017inversion}, but still require full PDE solves per sample and gradient-free MCMC. Physics-aware generative operators \cite{VADEBONCOEUR2023112369, zang2025dgno, seidman2023variational, zang2025psp, rahman2022generative, xia_bayesian_2022} can address forward and inverse problems jointly, though most lack physics-inspired architectures or demonstrations on inverse tasks. Diffusion models \cite{huang2024diffusionpde, yao2025guided, dasgupta_conditional_2025, LinHe2025Differentiable} have shown promise but often require large labeled datasets, rely on continuous relaxations with post hoc discretization, or employ guidance terms that do not guarantee consistent posterior sampling \cite{xu_rethinking_2025, lou_reflected_2023}.
A fundamental limitation shared by many methods is the lack of unified treatment: forward prediction and inverse inference are handled by separate models or objectives \cite{li2021physics}, despite stemming from the same physical system. This separation leads to inefficiencies, inconsistencies, and limited reuse of information.

In this work, we propose GenPANIS (Generative Physics-Aware Neural Implicit Solvers), a unified, physics-aware generative framework for solving both forward and inverse problems in multi-phase heterogeneous media. GenPANIS extends the discriminative PANIS framework \cite{Chatzop} from a task-specific predictor to a fully generative model capable of bidirectional inference within a single trained model. The generative formulation provides three key advantages: (i) principled uncertainty quantification through posterior sampling, (ii) ability to leverage abundant unlabeled microstructure data in addition to labeled pairs, and (iii) a learnable normalizing flow prior that mitigates posterior collapse and captures complex multimodal structure in latent space.

Our central design principle is to embed discrete material fields into a continuous latent space where differentiation and probabilistic inference are well defined, while preserving exact discrete microstructure configurations \cite{gomez-bombarelli_automatic_2018, wang2019coarse}. Rather than relaxing the material field itself \cite{LinHe2025Differentiable, huang2024diffusionpde, yao2025guided, li2021physics}, we introduce a latent variable representation enabling gradient-based inference without altering the underlying physics. Decoded microstructures are probabilistically sampled from discrete distributions (Bernoulli or categorical), ensuring physically admissible solutions without post-hoc corrections.

The latent representation couples to PDE solutions through an implicit, physics-aware decoder: a learned coarse-grained PDE model that preserves essential physics. Unlike black-box networks, this embedded solver maintains the mathematical structure of governing equations while remaining fully differentiable. The coarse-scale formulation serves as a physics-informed bottleneck, enabling accurate predictions with limited training data and extrapolation to varying microstructural statistics  or boundary conditions. Training uses physics residuals, eliminating reliance on expensive labeled datasets while supporting principled uncertainty quantification.
Crucially, the framework treats forward and inverse problems symmetrically through conditioning in latent space: forward prediction conditions on material fields, inverse inference conditions on PDE observations. Both are solved within the same probabilistic model through posterior inference, without retraining or architectural modification.

Section \ref{sec:Methodology} introduces the generative model architecture, data types, and training/prediction algorithms. Section \ref{sec:numerical} demonstrates the approach on inverse and forward problems governed by Darcy flow and Helmholtz equations in two-phase media, showing accurate recovery of discrete material fields from sparse or noisy observations, posterior uncertainty quantification, and generalization beyond training distributions. Section \ref{sec:conclusions} summarizes findings and outlines future directions.
\section{Methodology}
\label{sec:Methodology}

We consider a multiphase material system defined on a bounded domain $\Omega \subset \mathbb{R}^{n}$ governed by a partial differential equation (PDE) of the general form
\begin{equation}
    \mathcal{L}(u; c(\mathbf{s})) = f, \quad \forall \bs{s} \in \Omega,
\label{eq:pde}
\end{equation}
equipped with suitable boundary conditions. Here $u(\mathbf{s}) \in \mathcal{U}$ denotes the physical state (e.g., temperature,  pressure), $f$ is a source term, and $c(\mathbf{s})$ is a spatially varying material coefficient taking values in a finite set corresponding to distinct material phases, i.e. $c(\mathbf{s}) \in \{ c^{(0)},c^{(1)}, \ldots, c^{(P-1)}  \}$ where $P$ is the number of phases.

We use the term {\em microstructure} to refer to the spatial distribution of material phases over a spatial discretization into $d_x$ spatial cells $\Omega_j, j=1,2,\ldots,d_x$ (pixels in 2D or voxels in 3D),  which constitutes the primary unknown in the inverse problem. 
We represent the microstructure as a discrete-valued vector \cite{koutsourelakis_probabilistic_2006}
\be
\bx = \{x_j\}_{j=1}^{d_x}, \qquad x_j \in \{0,1,\dots,P-1\},
\label{eq:defx}
\ee
Each component $x_j$ accounts for the phase occupying cell $j$, and the corresponding PDE coefficient $c_j$ is
\begin{equation}
c(\bs{s}) = \sum_{j=1}^{d_x} c^{(x_j)} 1_{\Omega_j}(\bs{s})
\label{eq:xtoc}
\end{equation}
 In this way, the vector $\bx$ fully specifies the discrete coefficient field $c(\mathbf{s})$ in \refeq{eq:pde}, directly linking the microstructure to the PDE solution. Alternative representations, such as level-sets which model interfaces implicitly and are suitable for smooth boundaries \cite{osher1988fronts}, or graph-based encodings \cite{du2018microstructure, thomas2023materials} which operate on irregular meshes or network structures by treating spatial cells as graph nodes with the corresponding phase being a nodal feature, have been employed successfully for similar problems. We finally note that, in most practical cases, the dimension $d_x$ of $\bx$ must be very large to capture microstructural details and phase boundaries. The discretization of the PDE-solver (e.g., in a finite-volume/finite-element method) would generally need to be even finer to capture property variations, which implies a number of unknowns that is higher than $d_x$.

Given a noisy observation vector $\hat{\bs{u}}\in \mathbb{R}^{d_{\text{obs}}}$ of the physical state $u$:
\begin{equation}
    \hat{\bs{u}} = \mathbf{H}(u) + \boldsymbol{\eta},
    \label{eq:obsu}
\end{equation}
where $\boldsymbol{\eta}$ denotes additive noise, $\mathbf{H}: \mathcal{U} \to \mathbb{R}^{d_{\text{obs}}}$ is an observation operator that extracts measurements from the PDE solution $u$ (e.g., point evaluations, spatial averages, or linear functionals),  the goal is to identify the corresponding microstructure $\bx$. A Bayesian formulation places a prior $p(\bx)$ over admissible microstructures and defines the posterior distribution
\begin{equation}
    p(\bx \mid  \hat{\bs{u}}) \propto p( \hat{\bs{u}} \mid \bx) \, p(\bx),
    \label{eq:classicpost}
\end{equation}
where the likelihood $p(\hat{\bs{u}} \mid \bx)$ incorporates the PDE constraint via the forward map $\bx \overset{\text{Eq.\ref{eq:xtoc}}}{\mapsto} c
\overset{\text{Eq. \ref{eq:pde}}}{\mapsto} u
\overset{\text{Eq. \ref{eq:obsu}}}{\mapsto} \hat{\bs{u}}$. This framework is principled but presents the following challenges: (i) $\bx$ is a high-dimensional discrete variable, leading to a combinatorial posterior, (ii) posterior exploration requires repeated PDE solves, and (iii) gradients $\partial u / \partial \bx$ are not defined in the classical sense due to the piecewise-constant structure of $c(\mathbf{s})$. As a result, both sampling and optimization become difficult in such systems.

Our goal is to address these obstacles while maintaining the discrete  nature of the microstructure $\bx$. We introduce GenPANIS, a probabilistic latent-variable model  with the following key features:  
\bi
    \item The discrete microstructure $\bx$ is embedded into a continuous latent space $\bz$, which allows gradient-based optimization and probabilistic inference while preserving the discrete nature of the microstructure, unlike continuous relaxation methods that require post hoc corrections and can distort physical fidelity.
    \item A physics-aware decoder maps latent variables to {\em both} the PDE input $\bx$ and the (discretized) PDE solution $u$, disentangling their dependence and facilitating subsequent tasks.
    \item The generative model jointly represents $p(\bx,u)$, enabling both forward prediction (i.e. $u$ given $\bx$) and inverse inference (i.e. $\bx$ given (noisy) $u$) without retraining or separate architectures.
    \item Training is performed end-to-end in a single step, so that the latent embedding, PDE decoder, and predictive model are all optimized simultaneously with respect to the ultimate predictive objective. The framework can leverage labeled, unlabeled, and virtual data in a principled manner, and the loss arises naturally from the probabilistic formulation, avoiding ad-hoc relative weights.

\ei

This formulation amortizes the cost of PDE evaluations via an offline-trained, physics-aware surrogate that emulates {\em both} the forward and inverse maps. Once learned, the surrogate can be queried repeatedly to accelerate Bayesian (or deterministic) inverse problems without additional PDE solves, while still supporting uncertainty quantification in multiphase media. 
The remainder of this section presents the proposed framework in a structured manner. We begin with the latent-variable generative model (subsection \ref{sec:genMainBlocks}), followed by a discussion of the types of data that can be leveraged (subsection \ref{sec:dataTypes}), the training procedure based on stochastic variational inference (subsection \ref{sec:training}), and finally the posterior predictive framework for forward and inverse problems (subsection \ref{sec:prediction}). A summary of the key symbols used in the GenPANIS framework is contained in Table \ref{tab:notation}.

\begin{table}[!t]
\centering
\small
\setlength{\tabcolsep}{5pt}
\begin{tabular}{lp{11cm}}
\toprule
\textbf{Symbol} & \textbf{Definition / Meaning} \\
\midrule
$\mathbf{s}$ & Spatial location in problem domain $\Omega \subset \mathbb{R}^n$ \\
$u(\mathbf{s})$ & Physical state/PDE solution (e.g., temperature, pressure) \\
$c(\mathbf{s})$ & Spatially varying material coefficient taking discrete values $c^{(0)}, \dots, c^{(P-1)}$ \\
$\bx = \{x_j\}_{j=1}^{d_x}$ & Discrete microstructure vector with $x_j \in \{0,1,\dots,P-1\}$ (e.g., $x_j \in \{0,1\}$ for two-phase) \\
$\mathbf{y} \in \mathbb{R}^{d_\mathbf{y}}$ & PDE-solution coefficient vector (basis expansion coefficients or NN weights) \\
$\hat{\mathbf{u}}$ & Observed (noisy) PDE measurements, $\hat{\mathbf{u}} = \mathbf{H}(u) + \textrm{noise}$ \\
$\mathbf{z} \in \mathbb{R}^{d_z}$ & Latent variable vector (continuous embedding of $\bx$ and $\by$) \\
$p_\theta(\mathbf{z})$ & Learnable prior over latent variables (normalizing flow) \\
$p_\theta(\mathbf{x}|\mathbf{z})$ & Decoder for microstructure (logistic for $P=2$, categorical for $P>2$) \\
$p_\theta(\mathbf{y}|\mathbf{z})$ & Decoder for PDE solution coefficients $\by$, maps $\mathbf{z} \to \mathbf{y}$ \\
$\mathcal{R}(\mathbf{y})$ & Reconstruction map from coefficient vector $\mathbf{y}$ to function $u$, i.e., $u = \mathcal{R}(\mathbf{y})$ \\
$h(\cdot)$ & Lifting operator mapping coarse solution $\byy$ to PDE coefficients $\mathbf{y}$ \\
$\byy = \byy(\bxx)$ & Coarse PDE solution as function of coarse-grained PDE input $\bxx$ \\
$\bt = \{\bt_p, \bt_x, \bt_f, \boldsymbol{\sigma}_y \}$ & All trainable model parameters (prior and decoder parameters) \\
$\mathcal{D}_u$ & Unlabeled dataset (microstructures only) \\
$\mathcal{D}_v$ & Virtual dataset (microstructures + PDE residuals) \\
$\mathcal{D}_l$ & Labeled dataset (microstructure + PDE-solution pairs) \\
$p_\theta(\mathcal{D}|\theta)$ & Total marginal likelihood (product over all data types) \\
$q_\xi(\mathbf{z})$ & Variational/encoder distribution, amortized Gaussian $\mathcal{N}(g_\xi(\hat{\mathbf{x}}), \sigma_e^2 I)$ \\
$\xi = \{\xi_g, \sigma_e^2\}$ & Encoder parameters (NN parameters + encoder variance) \\
$r_{w_k}(\mathbf{x}, \mathbf{y})$ & Weighted PDE residual with weight function $w_k$ (for virtual observations) \\
\bottomrule
\end{tabular}
\caption{Key symbols used in the GenPANIS framework. Symbols with a hat $\hat{\cdot}$ indicate observed or given quantities.}
\label{tab:notation}
\end{table}

\subsection{Generative Model Architecture}
\label{sec:genMainBlocks}

\begin{figure}[t]
    \centering
    \includegraphics[width=0.7\linewidth]{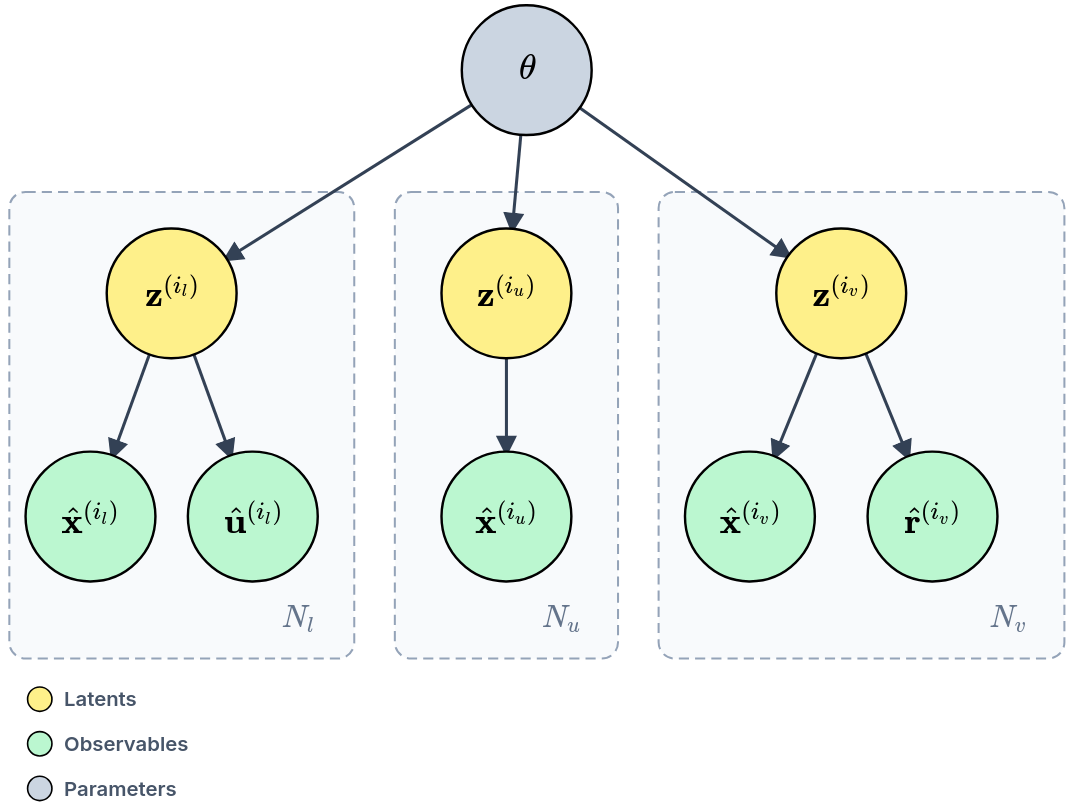}
    \caption{Probabilistic graphical model illustration for the GenPANIS framework. Plates indicate different data types (labeled, unlabeled, virtual). See text for details on model structure and parameterization. }
    \label{fig:mainBlocks}
\end{figure}

Our framework introduces a probabilistic latent-variable model that enables gradient-based inference on discrete microstructures. The model (Figure \ref{fig:mainBlocks}) consists of three core components: (i) a learnable prior $p_{\boldsymbol{\theta}}(\bz)$ over latent variables $\bz \in \mathbb{R}^{d_z}$ that encode admissible microstructures and their physical responses, (ii) a probabilistic decoder $p_{\boldsymbol{\theta}}(\bx \mid \bz)$ for discrete microstructures, and (iii) a physics-aware decoder $p_{\boldsymbol{\theta}}(u \mid \bz)$ that maps latent variables to PDE solutions. Together, these components generate coupled solution pairs $(\bx, u)$ while maintaining a structured, disentangled representation of their relationship.

\subsubsection{Learnable prior via normalizing flows.}
\label{sec:priorz}
We define the prior $p_{\boldsymbol{\theta}}(\bz)$ using a RealNVP normalizing flow \cite{dinh2016density} rather than a standard Gaussian. This choice mitigates auto-decoding \cite{alemi2018fixing} by increasing mutual information between data and latent variables, captures meaningful global structure (modes, clusters, manifolds) in latent space, and provides differentiable transformations essential for gradient-based inverse inference. The flow constructs $\bz$ through a sequence of $K$ invertible affine coupling transformations:
\begin{equation}
\bz = f_{\boldsymbol{\theta}_p}(\boldsymbol{\zeta}) = f_K \circ f_{K-1} \circ \cdots \circ f_1(\boldsymbol{\zeta}),
\end{equation}
where $\boldsymbol{\zeta} \sim \mathcal{N}(\mathbf{0}, \mathbf{I}_{d_z})$. By the change-of-variables formula:
\begin{equation}
\log p_{\boldsymbol{\theta}_p}(\bz)
=
\log p_0(\boldsymbol{\zeta})
+
\sum_{k=1}^{K}
\log
\left|
\det
\frac{\partial f_k^{-1}}{\partial \bz_k}
\right|,
\qquad
\boldsymbol{\zeta} = f_{\boldsymbol{\theta}}^{-1}(\bz).
\label{eq:priorz}
\end{equation}
Details regarding the total parameters $\bt_p$ induced by the prior and the network architecture are provided in \ref{appendix:shallowCNN}.

\subsubsection{Decoder for the PDE-input $p_{\boldsymbol{\theta}}(\bx \mid \bz)$}
\label{sec:decodex}
We adopt a low-capacity decoder to shift representational burden to the latent space. For binary microstructures ($P=2$), we use logistic PCA \cite{tipping_probabilistic_1998}:
\begin{equation}
    p_{\boldsymbol{\theta}}(\bx \mid \bz) = \prod_{j=1}^{d_x} 
    \sigma\!\big(a_j(\bz)\big)^{x_j} 
    \left(1 - \sigma\!\big(a_j(\bz)\big)\right)^{1-x_j},
    \label{eq:xrep}
\end{equation}
where $\sigma(\cdot)$ is the logistic sigmoid, $\mathbf{a}(\bz) = \mathbf{W}\bz + \mathbf{b}$, and $\boldsymbol{\theta}_{x} = \{\mathbf{W},\, \mathbf{b}\}$. For multiphase media ($P>2$), the Bernoulli factors are replaced by categorical distributions with softmax activation.

\subsubsection{The decoder for the PDE-output $p_{\bt}(u | \bz)$}
\label{sec:ydecoder}
We represent the PDE solution $u(\bs{s})$ via coefficient vector $\by \in \mathbb{R}^{d_{\by}}$ through a reconstruction map $\mathcal{R}: \mathbb{R}^{d_{\by}} \to \mathcal{U}$ such that $u = \mathcal{R}(\by)$. The vector $\by$ may correspond to basis function coefficients (e.g., Fourier, wavelet, finite-element) or neural network weights (e.g., DeepONet \cite{lu2021deeponet}, FNO \cite{li2020fourier}). Following \cite{Chatzop}, we employ a coarse-grained PDE embedding where latent variables $\bz$ first map to effective properties $\bxx$ via neural network $f_{\bt_f}(\bz)$ (see \ref{appendix:shallowCNN}), which are then mapped through the coarse solution operator $\byy(\bxx)$ and lifted to fine-scale coefficients $\by$ via operator $h(\byy)$ (details can be found in \cite{Chatzop}):
\begin{equation}
p_{\bt}(\by|\bz) = \delta \left(\by - h\big(\byy(f_{\bt_f}(\bz))\big) \right).
\label{eq:decodey}
\end{equation}
This lightweight physics-based bottleneck captures essential governing equation structure, enabling accurate predictions under varying boundary conditions even with limited training data \cite{grigo2019physics,Chatzop}.
Combined with the reconstruction map, this yields:
\begin{equation}
p_{\bt}(u|\bz) = \delta \left(u - \mathcal{R}\big(h(\byy(f_{\bt_f}(\bz)))\big) \right).
\label{eq:decodeu}
\end{equation}
The deterministic decoder facilitates computational efficiency during training, while the overall probabilistic framework preserves uncertainty quantification through the latent variable $\bz$.

The complete set of trainable model parameters is:
\begin{equation}
\bt = \{ \bt_p, \bt_{x}, \bt_f, \bs{\sigma}_{\by} \},
\label{eq:thetadef}
\end{equation}
comprising flow parameters, decoder networks, and observation noise variances.

\subsection{Data types and Model Likelihoods}
\label{sec:dataTypes}

A key advantage of the proposed generative framework is its ability to incorporate multiple forms of data during training. We formalize three data types and their corresponding likelihoods.

\subsubsection{Unlabeled data}

Unlabeled microstructures $\mathcal{D}_u = \{ \hat{\bx}^{(i_u)} \}_{i_u=1}^{N_u}$ (i.e. just microstructures) constitute the least expensive data modality. The likelihood of each sample is:
\begin{equation} 
    p_{\boldsymbol{\theta}}(\hat{\bx}^{(i_u)}) 
    = \int p_{\boldsymbol{\theta}}(\hat{\bx}^{(i_u)} \mid \bz^{(i_u)}) \, 
      p_{\boldsymbol{\theta}}(\bz^{(i_u)}) \, d\bz^{(i_u)},
\label{eq:LikeU} 
\end{equation}
with marginal likelihood 
\begin{equation}
    p(\mathcal{D}_u|\bt) = \prod_{i_u=1}^{N_u} p_{\bt}(\hat{\bx}^{(i_u)})
    \label{eq:marglikeu}
\end{equation}

\subsubsection{Virtual data}
Virtual observables \cite{kaltenbach2020incorporating,rixner2021probabilistic,Chatzop,scholz2025weak,zang2025dgno} encode the constraint that PDE residuals should vanish. Given $K$ weight functions $\{w_k \}_{k=1}^K$ and their weighted residuals $r_{w_k}(\bx, \by)$, we define the virtual observation vector $\hat{\mathbf{r}}^{(i_v)} = \{\hat{r}_{w_k}^{(i_v)}\}_{k=1}^K$ and treat zero residuals as pseudo-observations with noise model:
\begin{equation}
    0=\hat{r}_{w_k}^{(i_v)} = r_{w_k}(\hat{\bx}^{(i_v)}, \by^{(i_v)}) + \lambda^{-\frac{1}{2}} \epsilon_r^{(i_v)}, \quad \epsilon_r^{(i_v)} \sim \mathcal{N}(0, 1), \quad k=1,\ldots,K,
\end{equation}
yielding the virtual likelihood:
\begin{equation}
   p(\hat{\mathbf{r}}^{(i_v)}=\bs{0} |\hat{\bx}^{(i_v)}, \by^{(i_v)}, \bt) = \prod_{k=1}^{K} \sqrt{\frac{\lambda}{2\pi}} \exp\Big[-\frac{\lambda}{2}  r_{w_k}^2(\hat{\bx}^{(i_v)}, \by^{(i_v)}) \Big],
   \label{eq:vodef}
\end{equation}
where $\lambda$ controls enforcement strength\footnote{The specific choice of weight functions $w_k$ and basis functions for representing $u$ are detailed in Appendix~\ref{appendix:trialSolutions}.}. For $N_v$ pairs $\mathcal{D}_v = \{\hat{\bx}^{(i_v)}, \hat{\mathbf{r}}^{(i_v)}=\bs{0}\}_{i_v=1}^{N_v}$, combining \eqref{eq:vodef} with the decoder \eqref{eq:decodey} yields:
\begin{equation}
p_{\bt}(\hat{\bx}^{(i_v)}, \hat{\mathbf{r}}^{(i_v)}=\bs{0}) = \int p_{\bt}\Big(\hat{\mathbf{r}}^{(i_v)}=\bs{0} \,\big|\, h(\byy(f_{\bt_f}(\bz^{(i_v)}))), \hat{\bx}^{(i_v)}\Big) \, p_{\bt}(\hat{\bx}^{(i_v)}| \bz^{(i_v)}) \, p_{\bt}(\bz^{(i_v)}) ~d\bz^{(i_v)},
\label{eq:LikeO}
\end{equation}
with marginal likelihood 
\begin{equation}
    p(\mathcal{D}_v|\bt) = \prod_{i_v=1}^{N_v} p_{\bt}(\hat{\bx}^{(i_v)}, \hat{\mathbf{r}}^{(i_v)}=\bs{0}),
    \label{eq:marglikev}
\end{equation}

\subsubsection{Labeled data}

Labeled datasets $\mathcal{D}_l = \{\hat{\bx}^{(i_l)}, \hat{\mathbf{u}}^{(i_l)}\}_{i_l=1}^{N_l}$ consist of PDE input-output pairs obtained experimentally or via a reference solver. We model potentially  noisy observations via $p(\hat{\mathbf{u}}^{(i_l)} | u^{(i_l)})= \mathcal{N}(\hat{\mathbf{u}}^{(i_l)} | \bs{u}^{(i_l)},\text{diag}(\bs{\sigma}_{\by}))$, where $\bs{u}^{(i_l)}=\bs{B}~\by^{(i_l)}$ extracts values at observation points and $\bs{\sigma}_{\by}$ is learnable. Combined with \eqref{eq:decodeu}, the joint likelihood is:
\begin{equation}
    p_{\boldsymbol{\theta}}(\hat{\bx}^{(i_l)}, \hat{\bs{u}}^{(i_l)})  
    =\int \underbrace{ 
    \mathcal{N}(\hat{\mathbf{u}}^{(i_l)} | \bs{B} h\left(\byy(f_{\bt_f}(\bz))\right),\text{diag}(\bs{\sigma}_{\by}) )}_{p_{\boldsymbol{\theta}}(\hat{\bs{u}}^{(i_l)} \mid \bz^{(i_l)})}
    p_{\boldsymbol{\theta}}(\hat{\bx}^{(i_l)} \mid \bz^{(i_l)}) \, p_{\boldsymbol{\theta}}(\bz^{(i_l)}) \,  d\bz^{(i_l)},
\label{eq:LikeL}
\end{equation}
with marginal likelihood 

\begin{equation}
      p_{\bt}(\mathcal{D}_l | \bt) = \prod_{i_l=1}^{N_l} p_{\bt}(\hat{\bx}^{(i_l)}, \hat{\bs{u}}^{(i_l)})
      \label{eq:marglikel}
\end{equation}
We note finally that  acquiring a diverse labeled dataset of sufficient size is often computationally prohibitive; therefore, the use of labeled data should be limited whenever possible (i.e. $N_l$ should be preferably small or even zero).

\noindent The total training dataset is $\mathcal{D} = \{\mathcal{D}_u, \mathcal{D}_l, \mathcal{D}_v\}$, with all latent variables collected in $\mathcal{Z} = \{ \{ \bz^{(i_u)}\}_{i_u=1}^{N_u}, \{ \bz^{(i_l)}\}_{i_l=1}^{N_l}, \{ \bz^{(i_v)}\}_{i_v=1}^{N_v}\}$.

\subsection{Training with Stochastic Variational Inference (SVI)}
\label{sec:training}

Training the proposed generative model involves estimating the parameters $\bt$ by maximizing the marginal likelihood of the available data. Since the model includes latent variables $\mathcal{Z}$ for each data point, direct maximization is intractable. We adopt \emph{stochastic variational inference (SVI)} \cite{hoffman2013stochastic} to overcome this challenge.

Combining all data types, the total marginal likelihood is:
\begin{equation}
p(\mathcal{D}|\bt) = \underbrace{\prod_{i_u=1}^{N_u} p_{\bt}(\hat{\bx}^{(i_u)})}_{\text{unlabeled data}}
                       \underbrace{\prod_{i_l=1}^{N_l} p_{\bt}(\hat{\bx}^{(i_l)}, \hat{\bs{u}}^{(i_l)})}_{\text{labeled data}}
                       \underbrace{\prod_{i_v=1}^{N_v} p_{\bt}(\hat{\bx}^{(i_v)}, \hat{\mathbf{r}}^{(i_v)})}_{\text{virtual data}}.
\label{eq:LikeMarg}
\end{equation}
We obtain maximum likelihood estimates of $\bt$ by maximizing $p(\mathcal{D} | \bt)$\footnote{In all simulations, a very vague prior on $\bt$ of the form $\mathcal{N}\left(\bs{0}, \sigma_{\bt}^2 \bs{I} \right)$ with variance $\sigma_{\bt}^2 = 10^{16}$ was employed to avoid potential degeneracies. Hence, MAP rather than ML estimates were obtained, but as the prior's role is negligible, it is omitted from the presentation.}. A key advantage of this formulation is that it provides a rigorous probabilistic synthesis of terms corresponding to each data modality without requiring ad-hoc relative weights that need fine-tuning.

To handle the intractable marginal likelihood, we employ an amortized variational inference scheme \cite{kingma2015variational}. We introduce an auxiliary density $q_{\bksi}(\mathcal{Z})$ on the latent variables, parameterized by $\bksi$, which serves as an approximation to the model posterior. This yields a lower bound $\mathcal{F}(\bt, \bksi)$ on the marginal log-likelihood (Evidence Lower Bound, or ELBO), which we maximize iteratively with respect to both $\bt$ and $\bksi$ until convergence.

We employ a factorized variational distribution shared across all data types:
\begin{equation}
    q_{\bksi}(\mathcal{Z}) = \prod_{i=1}^N q_{\bksi}(\bz^{(i)}) = \prod_{i=1}^N\mathcal{N}\big(\bz^{(i)} | g_{\bksi_g}(\hat{\bx}^{(i)}), \sigma_e^2 I \big), \quad i=1,\dots,N=N_u+N_l+N_v,
\label{eq:encoder}
\end{equation}
where $\hat{\bx}^{(i)}$ denotes the associated microstructure in each case (see section \ref{sec:dataTypes}). Amortization is achieved via the neural network $g_{\bksi_g}$ (details in \ref{appendix:shallowCNN}), which maps each observed microstructure to latent space parameters. Despite the simple isotropic covariance structure, these probabilistic encoders have been shown to yield better-regularized models than deterministic alternatives \cite{alemi2018fixing}, which we also confirmed experimentally. The complete variational parameter vector is:
\begin{equation}
    \bksi = \left\{ \bksi_g, \sigma_e^2 \right\}.
\end{equation}

The ELBO $\mathcal{F}(\bt,\bksi)$ consists of contributions from each data modality:
\be
\mathcal{F}(\bt,\bksi) = \mathcal{F}_u(\bt,\bksi)+\mathcal{F}_v(\bt,\bksi)+\mathcal{F}_l(\bt,\bksi),
\label{eq:elbo}
\ee
which arises by applying Jensen's inequality to each marginal likelihood. For {\em unlabeled} data (\refeq{eq:marglikeu}):
\be
\log p(\mathcal{D}_u|\bt) \ge \sum_{i_u=1}^{N_u} \left\langle \log \frac{ p_{\boldsymbol{\theta}}(\hat{\bx}^{(i_u)} \mid \bz^{(i_u)}) \, 
      p_{\boldsymbol{\theta}}(\bz^{(i_u)}) }{ q_{\bksi}(\bz^{(i_u)})} \right\rangle_{q_{\bksi}(\bz^{(i_u)})} =\mathcal{F}_u(\bt,\bksi),
\label{eq:elbou}
\ee
for {\em virtual} data (\refeq{eq:marglikev}):
\be
\small
\log p(\mathcal{D}_v|\bt) \ge \sum_{i_v=1}^{N_v} \left\langle \log \frac{ p_{\bt}\Big(\hat{\mathbf{r}}^{(i_v)}=0 \,\big|\, h(\byy(f_{\bt_f}(\bz^{(i_v)}))), \hat{\bx}^{(i_v)}\Big) \, p_{\bt}(\hat{\bx}^{(i_v)}| \bz^{(i_v)}) \, p_{\bt}(\bz^{(i_v)}) }{ q_{\bksi}(\bz^{(i_v)})} \right\rangle_{q_{\bksi}(\bz^{(i_v)})} =\mathcal{F}_v(\bt,\bksi),
\label{eq:elbov}
\ee
and for {\em labeled} data (\refeq{eq:marglikel}):
\be
\log p(\mathcal{D}_l|\bt) \ge \sum_{i_l=1}^{N_l} \left\langle \log \frac{ p_{\boldsymbol{\theta}}(\hat{\bs{u}}^{(i_l)} \mid \bz^{(i_l)}) \, p_{\boldsymbol{\theta}}(\hat{\bx}^{(i_l)} \mid \bz^{(i_l)}) \, p_{\boldsymbol{\theta}}(\bz^{(i_l)}) }{ q_{\bksi}(\bz^{(i_l)})} \right\rangle_{q_{\bksi}(\bz^{(i_l)})} =\mathcal{F}_l(\bt,\bksi).
\label{eq:elbol}
\ee
If any data type is unavailable, the corresponding term is omitted from $\mathcal{F}$.

We maximize $\mathcal{F}(\bt, \bksi)$ using stochastic gradient ascent with the ADAM optimizer \cite{kingma2014adam}. Gradients are approximated via Monte Carlo estimates and the reparameterization trick \cite{kingma2015variational}.
The training procedure is summarized in Algorithm \ref{alg:training}.

\begin{algorithm}[t]
\begin{algorithmic}[1]
    \State  Initialize $\bt\gets \bt_0$; $\bksi\gets \bksi_0$; $\ell \gets 0$
    \While{$\mathcal{F}(\bt, \bksi)$ not converged} 
        \State Draw samples $\mathcal{Z}^{(\ell)} \sim q_{\bksi}(\mathcal{Z})$ using reparameterization trick
        \State Approximate $\mathcal{F}(\bt_{\ell}, \bksi_{\ell}) \approx \hat{\mathcal{F}}(\bt_{\ell}, \bksi_{\ell})$ via Monte Carlo
        \State Compute $ \{\nabla_{\bt} \hat{\mathcal{F}}(\bt_{\ell}, \bksi_{\ell}), \nabla_{\bksi} \hat{\mathcal{F}}(\bt_{\ell}, \bksi_{\ell}) \}$ via backpropagation
        \State $ \{ \bt_{\ell+1}, \bksi_{\ell+1} \} \gets $ ADAM update using $ \{ \nabla_{\bt} \hat{\mathcal{F}}(\bt_{\ell}, \bksi_{\ell}), \nabla_{\bksi} \hat{\mathcal{F}}(\bt_{\ell}, \bksi_{\ell}) \}$
        \State $\ell \gets \ell + 1$
    \EndWhile
\end{algorithmic}
\caption{Training via Stochastic Variational Inference}
\label{alg:training}
\end{algorithm}

\subsection{Predictive Estimates for Forward and Inverse Problems}
\label{sec:prediction}

After training, the model defines a joint probabilistic generative model over the latent variable $\bz$, the PDE-input $\bx$, and the PDE-output $u$. This enables the trained model to serve as a surrogate for both data generation and prediction: synthetic samples can be drawn without solving the PDE, and conditional samples can be generated to address forward or inverse problems. The key enabling feature is the mediating role of $\bz$, which disentangles the input-output relationship and supports efficient conditional inference.

\subsubsection{Inverse Problems}
\label{sec:predinverse}

For an inverse problem as in \eqref{eq:obsu}, the probabilistic predictive estimate for the microstructure $\bx$ given observations $\hat{\bs{u}}$ is:
\be
\begin{array}{ll}
p_{\bt}(\bx | \hat{\bs{u}}) & \propto p_{\bt}(\bx , \hat{\bs{u}}) \\
& = \int p_{\bt}(\bx |\bz) ~p_{\bt}(\bz|\hat{\bs{u}}) ~d\bz,
\end{array}
\label{eq:condxu}
\ee
where we have marginalized over the PDE solution $\bs{u}$ and applied Bayes' rule. Critically, inferring $\bx$ does \emph{not} require search or sampling in the high-dimensional, discrete microstructure space. Instead, the problem reduces to two steps: (i) infer $\bz$ from $p_{\bt}(\bz|\hat{\bs{u}})$ in the low-dimensional continuous latent space, and (ii) propagate $\bz$ through the decoder $p_{\bt}(\bx |\bz)$ to obtain discrete microstructure samples.

This constitutes a central practical advantage: the posterior $p_{\bt}(\bz|\hat{\bs{u}})$ is defined over a continuous latent space, enabling gradient-based probabilistic inference. Derivatives of $\log p_{\bt}(\bz|\hat{\bs{u}})$ are readily available through backpropagation, allowing deployment of modern inference techniques such as Hamiltonian Monte Carlo \cite{duane1987hybrid}. Computational efficiency is preserved since gradients only flow through neural networks and the coarse-grained PDE model of section~\ref{sec:ydecoder}, requiring no fine-scale PDE solves.

Assuming additive Gaussian noise in \eqref{eq:obsu} with variance $\sigma_n^2$, we have
$p(\hat{\bs{u}} | \bs{u})=\mathcal{N}( \hat{\bs{u}} | \bs{C} \bs{y}, \sigma_n^2\bs{I})$,
where matrix $\bs{C}$ extracts PDE-outputs at sensor locations. Combined with \eqref{eq:decodeu}, the tractable posterior is:
\be
p_{\bt}(\bz|\hat{\bs{u}}) \propto \mathcal{N} \left( \hat{\bs{u}}~| \bs{C}~ h\left(\byy(f_{\bt_f}(\bz))\right), \sigma_n^2\bs{I}\right) ~p_{\bt}(\bz).
\label{eq:noisyInvProb}
\ee
In practice, inference is performed  as detailed in Algorithm \ref{alg:inference}. The computational cost is dominated by  sampling in the low-dimensional latent space (typically $d_z \sim 50$) rather than evaluating expensive forward models on high-dimensional discrete fields ($d_x \sim 10^4$), and requires only gradients through neural networks and the coarse-grained PDE solver—no fine-scale PDE evaluations are needed.

\begin{algorithm}[!t]
\begin{algorithmic}[1]
    \Require Observations $\hat{\bs{u}}$, trained model parameters $\bt$,  number of samples $L$
    \Ensure Posterior samples $\{\bx^{(l)}\}_{l=1}^L$ and statistics $(\bar{\bx}, \text{Var}[\bx])$
    \State Initialize $\bz_0 \sim p_{\bt}(\bz)$ \Comment{Using trained prior \refeq{eq:priorz}}
    \For{$l = 1$ to $L$}
        \State Sample $\bz^{(l)} \sim p_{\bt}(\bz|\hat{\bs{u}})$ using HMC \Comment{Gradients via backpropagation through Eq.~\ref{eq:noisyInvProb}}
        \State Sample $\bx^{(l)} \sim p_{\bt}(\bx|\bz^{(l)})$ \Comment{Draw from Bernoulli decoder Eq.~\ref{eq:xrep}}
    \EndFor
    \State Compute $\bar{\bx} = \frac{1}{L}\sum_{l=1}^L \bx^{(l)}$ \Comment{Posterior mean}
    \State Compute $\text{Var}[\bx]_j = \frac{1}{L}\sum_{l=1}^L (\bx_j^{(l)} - \bar{\bx}_j)^2$
 \Comment{Element-wise posterior variance}
    \State \Return $\{\bx^{(l)}\}_{l=1}^L$, $\bar{\bx}$, $\text{Var}[\bx]$
\end{algorithmic}
\caption{Inverse inference via HMC in latent space}
\label{alg:inference}
\end{algorithm}

\subsubsection{Forward Problems}
\label{sec:predforward}

For a given microstructure $\hat{\bx}$, forward prediction follows an analogous procedure. The predictive distribution is:
\begin{equation}
p_{\bt}(u | \hat{\bx}) = \int p_{\bt}(u | \bz)  p_{\bt}(\bz| \hat{\bx})  d\bz,
\label{eq:forProb}
\end{equation}
where $p_{\bt}(\bz | \hat{\bx}) \propto p_{\bt}(\hat{\bx} | \bz) p_{\bt}(\bz)$ using the learned prior \eqref{eq:priorz} and decoder \eqref{eq:xrep}. Samples are drawn from $p_{\bt}(\bz|\hat{\bx})$ via HMC, then propagated through $p_{\bt}(u |\bz)$ in \eqref{eq:decodeu}. Alternatively, one could use the trained encoder $q_{\bksi}(\bz|\hat{\bx})$ in \refeq{eq:encoder} which approximates $p_{\bt}(\bz|\hat{\bx})$ but can be sampled much more easily.   Uncertainty in the forward prediction arises naturally from the posterior variance of $\bz$ given $\hat{\bx}$, capturing both aleatoric uncertainty (e.g., observation noise during training) and epistemic uncertainty (model uncertainty about the latent representation).

The latent-variable formulation directly addresses the three challenges identified in Section~\ref{sec:Methodology}: the combinatorial posterior over discrete $\bx$ becomes a continuous posterior over $\bz$ enabling gradient-based inference; the expensive PDE map is replaced by a lightweight surrogate eliminating repeated solves; and non-differentiability is resolved by working in latent space where $\nabla_{\bz} u$ is well-defined. Moreover, the same trained model handles both forward and inverse problems through conditioning, and the embedded coarse-grained solver enables extrapolation to unseen boundary conditions and microstructural statistics.

\subsubsection{Evaluation Metrics}

We assess performance using a test dataset $\mathcal{D}_t = \{ \hat{\bx}^{(i_t)}, \hat{\bs{u}}^{(i_t)} \}_{i_t=1}^{N_t}$ consisting of $N_t$ PDE input-output pairs obtained using a conventional FEM solver on a fine mesh.

\noindent \textbf{Relative $L_2$ Error $\epsilon$ (forward problems).}
Given reference pairs $\mathcal{D}_t$ and $\bar{\bs{u}}^{(i_t)}$ the posterior mean computed from samples drawn from $p_{\bt}(\bs{u} |\hat{\bx}^{(i_t)})$, we define: 
\begin{equation}
\epsilon = \frac{1}{N_{t}} \sum_{i_t=1}^{N_{t}} \frac{ \| \hat{\bs{u}}^{(i_t)} - \bar{\bs{u}}^{(i_t)} \|_2 }{ \| \hat{\bs{u}}^{(i_t)} \|_2}.
\label{eq:errorf}
\end{equation}

\noindent \textbf{Pixel-wise Accuracy $\mathbf{PA}$ (inverse problems).}
For a synthetic inverse problem based on a single instance from $\mathcal{D}_t$ with reference microstructure $\hat{\bx}^{(i_t)}$ and corresponding solution $\hat{\bs{u}}^{(i_t)}$, we compute the posterior mean $\bar{\bs{x}}$ using samples from $p_{\bt}(\bx | \hat{\bs{u}}^{(i_t)})$ and threshold at 0.5 to compare with $\hat{\bx}^{(i_t)}$:
\begin{equation}
   \mathrm{PA} = \frac{1}{d_x}\sum_{j=1}^{d_x} \mathbf{1}\!\left(\mathcal{T}_{0.5}(\bar{\bx}_j)  = \hat{\bx}^{(i_t)}_j\right),
\end{equation}
where $\mathcal{T}_{0.5}(\bar{x}_j)=1$ if $\bar{x}_j>0.5$ and $0$ otherwise. This metric \cite{yao2025guided} measures the proportion of cells matching the ground truth.

\section{Numerical Illustrations}
\label{sec:numerical}
This section presents a suite of numerical experiments demonstrating the performance of GenPANIS on two different PDEs. Results are compared against the Physics-Informed Neural Operator (PINO) \cite{li2021physics} and FunDPS \cite{yao2025guided}, a guided diffusion model on function spaces. PINO has been widely used as a benchmark in prior studies \cite{huang2024diffusionpde} and offers point estimates  for both forward and inverse problems. FunDPS represents one of the most recent and computationally intensive models for PDEs, achieving, to the best of our knowledge, state-of-the-art results for challenging inverse problems with sparse or noisy observations.
We note that both methods employ post-hoc thresholding to produce discrete microstructure estimates in inverse problems, along with additional regularization as detailed in the respective papers. In contrast, GenPANIS directly defines a distribution over discrete-valued $\bx$, eliminating the need for such corrections while preserving exact discrete configurations throughout inference.

In the following subsections, we aim:

\bi
\item to assess the ability of GenPANIS to extrapolate reliably in out-of-distribution scenarios, including cases with sparse or highly noisy observations, which additionally have drastically different values than the ones during training.

\item to compare the accuracy of posterior approximation with the aforementioned state-of-the-art methods, i.e. PINO \cite{li2021physics} and FunDPS \cite{yao2025guided} under limited or noisy data.

\item to demonstrate that lightweight, physics-aware models like GenPANIS can achieve comparable or superior performance, especially in extrapolative cases, compared to models with orders of magnitude more parameters. 

\item to illustrate the benefits of leveraging a mixture of data types, particularly the least-expensive  unlabeled data and virtual observables, to improve learning efficiency and predictive accuracy.
\ei

The implementation of GenPANIS is based fully on PyTorch \cite{paszke2019pytorch}. All the models were trained on an NVIDIA RTX 4090 GPU. Some indicative training times are shown in Table \ref{tab:modelStats}. All reference solutions were obtained using Fenics \cite{fenics2015} with a uniform, finite element mesh consisting of 16641 nodes. A GitHub repository containing the associated code and illustrative examples will become available upon publication at \href{https://github.com/pkmtum/GenPANIS}{https://github.com/pkmtum/GenPANIS}.

\begin{table}[!t]
\centering
\begin{tabular}{|c|c|c|c|}
\hline
Model & PINO & FunDPS & GenPANIS \\
\hline
Training time (in $hrs$)   & $38.2$ & $44.5$ & $\mathbf{0.6}$ \\
\hline
Total Parameters (in millions)   & $13.1$ & $183.3$ & $\mathbf{2.9}$ \\
\hline
\end{tabular}
\caption{Training times and total number of trainable parameters for each model when training using $10000$ labeled data (Darcy Flow Equation). In contrast to PINO and GenPANIS, FunDPS can utilize only labeled data during training. This is why we train all models exclusively with labeled data for results where comparisons are feasible.}
\label{tab:modelStats} 
\end{table}

\subsection{Governing Equations}

We consider the following two elliptic PDEs:
\bi
\item  \textbf{Inhomogeneous Darcy Flow Equation:} Most of the state-of-the-art models \cite{yao2025guided, li2021physics, huang2024diffusionpde} have considered the following Darcy flow problem:
\begin{equation}
\begin{array}{lll}
 \nabla \cdot \left( - c(\bs{s}; \bx) \nabla  u(\bs{s})\right) = f, ~~~ \mathbf{s} \in \Omega = \left[0, ~ 1 \right] \times\left[0, ~ 1 \right] \\
u(\bs{s}) = u_0, ~~~ \mathbf{s} \in \Gamma_D,
\end{array}
\label{eq:Darcy}
\end{equation}
where $c(\bs{s}, \bx)$ is a permeability field parameterized by the vector $\bx$ (more details in the sequel), $f$ is the source term, and $u_0$ is the Dirichlet boundary condition. If not stated otherwise, $f=100$ and $u_0=0$ in the following experiments.
\item  \textbf{Propagation of acoustic waves (Helmholtz Equation)}: We consider the following  Helmholtz equation in the frequency domain:

\begin{equation}
\begin{array}{lll}
\nabla^2 u(\bs{s}) + \omega^2 c(\bs{s}; \bx) u(\bs{s}) 
= 0, ~~~ \bs{s} \in \Omega = \left[0, ~ 1 \right] \times \left[0, ~ 1 \right] \\
u(\bs{s}) = 1, ~~~ \bs{s} \in \Gamma_D.
\end{array}
\label{eq:Helmholtz}
\end{equation}

In the equation above, $u(\bs{s})$ is the pressure field and $c(\bs{s}; \bx)$  the reciprocal of the squared wave speed, which depends on the material microstructure $\bx$. In  all the subsequent illustrations,  the  angular frequency  was $\omega=1 ~rad/s$.
We note that while some recent models have considered this equation \cite{huang2024diffusionpde, yao2025guided}, they have typically used the source term as the input. In that case, the solution depends linearly on the right-hand side, making the mapping relatively easy to approximate. In contrast, the present work focuses on the more challenging, nonlinear dependence of the solution on the microstructure 
$\bx$, which is central to inverse heterogeneous media problems.
\ei

\subsection{Data Generation}

The material property fields in the equations above  are obtained by thresholding a zero-mean Gaussian process with a covariance function:
\begin{equation}
    k(\mathbf{s}, \mathbf{s}') = \exp\left( -\frac{\|\mathbf{s} - \mathbf{s}'\|^2}{l^2} \right),
\end{equation}
\noindent where the parameter $l$ governs the correlation length of the field.

For all the subsequent experiments, 
the dimension of the resulting binary microstructures is $d_{\bs{x}}=128^2=16384$.
Each pixel was assigned two possible values, namely $1$ and $\frac{1}{CR}$, where $CR$ indicates the contrast ratio between the two material phases and significantly influences the order of the microstructural statistics that affect the response. In the subsequent illustrations:
\bi
\item $CR=10$ for the  Darcy flow (\refeq{eq:Darcy}) and,
\item $CR=20$ for the Helmholtz equation (\refeq{eq:Helmholtz}).
\ei

\subsection{Training and Inference Details for the Compared Models}

Unless stated otherwise, all models are trained using  $10,000$ labeled data, i.e. $(\hat{\bx}, \hat{\bs{u}})$ pairs.  While GenPANIS does not require labeled data, comparison methods such as FunDPS do; using the same labeled dataset ensures a fair comparison. In Section \ref{sec:exp8}, we investigate how incorporating unlabeled and virtual data affects the accuracy of GenPANIS.

In order to carry out inference in the latent space $\bz$ as detailed in Algorithm \ref{alg:inference},  the hamiltorch Python package \cite{cobb2021scaling} was used. In the case of PINO, the inverse problems  were solved by applying the forward model method described in Section 3.4 of (\cite{li2021physics}, Equation (16))\footnote{We did not use the inverse model method presented therein  because it requires training an additional inverse operator, which doubles the cost and did not yield significant improvements in our experiments.}. 
For FunDPS,  inference was carried out using guided diffusion sampling, as described in \cite{huang2024diffusionpde, yao2025guided}.

\noindent \textbf{Coarse-Grained Implicit Solver:}
One of the key components of the proposed architecture is the coarse-grained implicit solver, described in detail in \cite{Chatzop}. Since this component is not novel, we only provide a brief overview and refer  the interested reader to \cite{Chatzop} for full details. The solver is defined on a regular  $16 \times 16$ grid with a total of $512$ triangular finite elements, each assumed to have a constant material property (e.g., permeability or wave speed) represented by the vector $\bxx$ of section \ref{sec:ydecoder} (i.e. $\dim(\bxx)=512 \ll \dim(\bx)=16384$). The discretized PDE solution at the nodal values is represented by the vector $\byy$ (i.e. $\dim(\byy)=256$).

\subsection{Assessment of model's components}
Before discussing model performance, we assess two aspects of the proposed model, namely the dimension of the latent vector $\bz$ and the form of the learnable prior $p_{\bt}(\bz)$, which have a significant impact on the model's size and training effort as well as on its accuracy.

\subsubsection{Determining $\dim(\mathbf{z})$}
\label{sec:determz}

We conduct a small empirical  study to determine the optimal $\dim(\bz)$  for both the forward and inverse problems, while simultaneously considering the total number of trainable parameters. The specific instances and observations for the forward problems are the same as in section~\ref{sec:forwardIllustrations}, and for the inverse problems, as in Section~\ref{sec:exp1}. The results of this study are presented in Table \ref{tab:paramDimz}, where we observe that the predictive accuracy for the forward problem increases as the dimension of $\bz$ increases. The accuracy of the inverse problem initially increases, reaching a peak around $\dim(\bz) \approx 60$, and then it begins to deteriorate very slightly. Furthermore, we observe that the model saturates quickly, achieving (near) peak performance already at $\dim(\bz)=30$. This property can be attributed to the coarse-grained model incorporated in the decoder $p_{\bt}(u|\bz)$ (section \ref{sec:ydecoder}) that serves as a physics-aware information bottleneck.
In the following illustrations, we use $\dim(\bz)=60$, which provides a good balance between accuracy and computational cost. Finally, we report  the number of training parameters in Table \ref{tab:modelStats}, which are one to two orders of magnitude fewer than those in PINO or FunDPS, which have a significant impact on training times, as one can observe.

\begin{table}[H]
\centering
\begin{tabular}{|c|c|c|c|}
\hline
$\dim(\bz)$ & $L_2$ Relative Error (Forward) & Pixel Accuracy (Inverse) & Total Parameters \\
\hline
20    & $8.37 \cdot 10^{-2}$ & $0.936$ & $\mathbf{1.3 \cdot 10^6}$  \\
\hline
30    & $5.91 \cdot 10^{-2}$ & $0.981$ & $1.7 \cdot 10^6$   \\
\hline
40    & $7.17 \cdot 10^{-2}$ & $\mathbf{0.985}$ & $2.1 \cdot 10^6$   \\
\hline
50    & $4.68 \cdot 10^{-2}$ & $0.976$ & $2.5 \cdot 10^6$   \\
\hline
60    & $4.33 \cdot 10^{-2}$  & $\mathbf{0.985}$ & $ 2.9 \cdot 10^6$ \\
\hline
70    & $7.7 \cdot 10^{-2}$  & $\mathbf{0.985}$  & $3.3 \cdot 10^6$  \\
\hline
100    & $\mathbf{3.3 \cdot 10^{-2}}$ & $0.976$ & $4.5 \cdot 10^6$    \\
\hline
150    & $\mathbf{3.3 \cdot 10^{-2}}$ & $0.975$ & $6.5 \cdot 10^6$   \\
\hline
200    & $4.7 \cdot 10^{-2}$ & $0.974$ & $ 8.5 \cdot 10^6$   \\
\hline
\end{tabular}
\caption{Relative error (forward problem), pixel accuracy (inverse problem), and total number of trainable parameters for different values of $\dim(\bz)$ in GenPANIS.}
\label{tab:paramDimz}
\end{table}

\subsubsection{The Effect of the Learnable Prior $p_{\bt}(\bz)$}

In this section, we demonstrate the importance of the learnable prior model $p_{\bt}(\bz)$ (Section \ref{sec:priorz}). Unlike a standard normal prior, the structured prior captures correlations in the latent space induced by the underlying physics, enabling more accurate predictions for both forward and inverse problems, particularly in regions of the parameter space far from the training distribution, as we show in the ensuing experiments. Table \ref{tab:flowVSstdNormal} compares results for both tasks between the standard normal and the proposed prior, confirming that the learnable prior consistently improves accuracy and robustness. 

\begin{table}[H]
\centering
\begin{tabular}{|c|c|c|}
\hline
Type of prior & $L_2$ Relative Error (Forward) & Pixel Accuracy (Inverse) \\
\hline
NVP flow model with 12 layers & $\mathbf{7.80 \cdot 10^{-2}}$ & $\mathbf{0.945}$    \\
\hline
$\mathcal{N}\left( \bs{0}, \bs{I}\right)$    & $2.26 \cdot 10^{-1}$ & $0.920$    \\
\hline
\end{tabular}
\caption{Results for the forward and inverse problems obtained when the flow model of section \ref{sec:priorz} is used as a prior for the latent variables $\bz$ vs. a standard  normal $\mathcal{N}\left( \bs{0}, \bs{I}\right)$.}
\label{tab:flowVSstdNormal}
\end{table}

\subsection{Forward Problems}
\label{sec:forwardIllustrations}
For completeness, we briefly examine forward prediction performance, i.e., computing the PDE solution $u$ given the input microstructure $\bx$ (Section~\ref{sec:predforward}). While forward prediction is an important capability, the primary contribution of GenPANIS lies in addressing inverse problems where the discrete nature of microstructures and partial observations pose fundamental challenges. We therefore provide representative forward results to establish baseline performance before focusing on the inverse setting.

\begin{figure}[H]
    \centering

    \begin{subfigure}{\textwidth}
        \centering
        \includegraphics[width=1\linewidth]{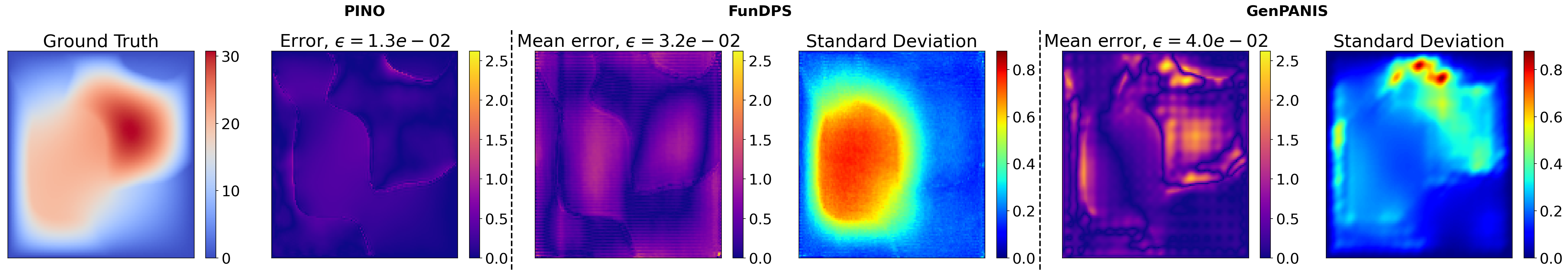}
        \caption{Predictions for the Darcy Flow Equation}
        \label{fig:exp1forwardD}
    \end{subfigure}

    \vspace{1em}

    \begin{subfigure}{\textwidth}
        \centering
        \includegraphics[width=1\linewidth]{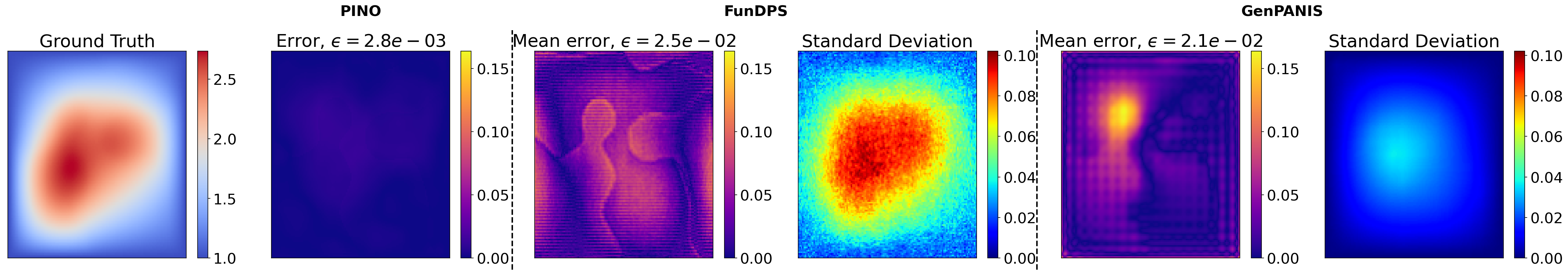}
        \caption{Predictions for the Helmholtz Equation}
        \label{fig:exp1forwardH}
    \end{subfigure}

    \vspace{1em}

    \caption{Forward problem predictions of the three models.}
    \label{fig:exp1_forward}
\end{figure}

Figure~\ref{fig:exp1_forward} shows posterior mean predictions (in terms of the error with the ground truth) and standard deviations for both PDEs. PINO provides deterministic point estimates and achieves the highest accuracy on both equations, benefiting from its direct forward operator architecture. Among the probabilistic models, FunDPS achieves slightly lower error than GenPANIS on the Darcy flow (Equation \ref{eq:Darcy}), while GenPANIS performs marginally better on the Helmholtz equation (\refeq{eq:Helmholtz}). Both models produce reasonable uncertainty estimates, with posterior standard deviations reflecting spatial variability in the solution fields. Notably, GenPANIS achieves competitive forward prediction accuracy despite having 4.5$\times$ fewer parameters than PINO and 63$\times$ fewer than FunDPS, demonstrating the efficiency of the physics-informed latent representation.

The remainder of this section focuses on inverse problems, where observations are partial, noisy, or sparse—conditions under which the advantages of GenPANIS become most pronounced.

\subsection{Inverse Problems}
We evaluate the proposed approach primarily on inverse problems, where challenges related to noise, sparsity of observations, and limited data are most pronounced. The experiments are organized to first consider {\em within-distribution} settings under progressively more restrictive observation regimes, followed by studies on the effect of training data availability. We then examine {\em out-of-distribution} scenarios to assess robustness with respect to changes in input statistics, boundary conditions, and microstructural properties. The section concludes with experiments illustrating the use of unlabeled data and virtual observables during training.

\subsubsection{Within-Distribution Inverse Problems }
We begin by examining inverse problems in a within-distribution setting, where the test instances are drawn from the same data-generating process as the training set. The goal of these experiments is to assess how each model handles increasing levels of uncertainty and decreasing information content in the observations, while keeping the underlying physics unchanged.

\paragraph{Full Observations for various levels of noise}
\label{sec:exp1}

We first consider inverse problems with full observations, i.e. on a regular $128 \times 128$ grid of the PDE-solution field and systematically vary the observation noise. This setting isolates the effect of noise on posterior recovery and allows for a direct comparison of the inferred distributions against a reference posterior obtained via sampling. 

\begin{figure}[H]
    \centering
    \includegraphics[width=1.0\linewidth]{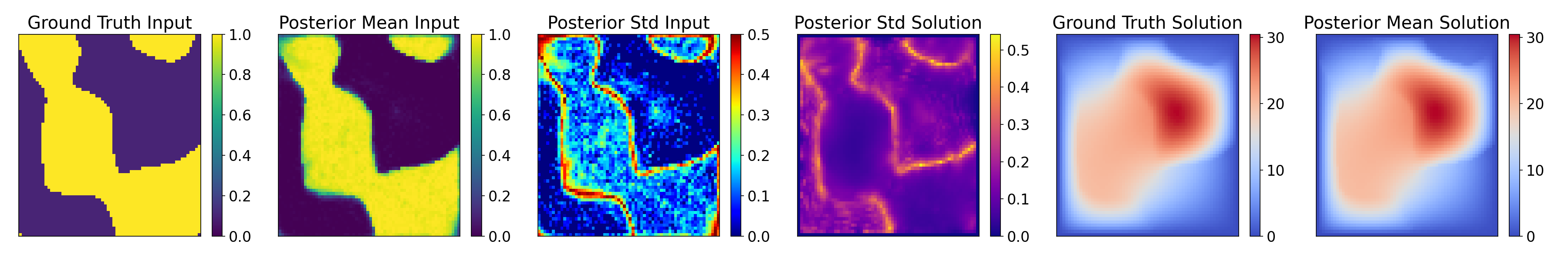}
    \caption{Reference posterior statistics for Darcy flow obtained from one million HMC samples. Full observations were used  with $SNR=100$. }
    \label{fig:posteriorExp1}
\end{figure}

In most related works \cite{huang2024diffusionpde, yao2025guided, li2021physics}, inverse problems with noisy observations are primarily assessed using point estimates. 
While informative, such summaries provide only a limited view of a model’s inferential capability. A more principled evaluation requires examining the full posterior, even if this can only be done in restricted settings.
For this reason, we compare the posterior distributions inferred by each model against the reference posterior (see \refeq{eq:classicpost}). For the low-noise case ($\mathrm{SNR}=100$), this reference posterior is shown in Figure~\ref{fig:posteriorExp1}. One million  samples from the reference posterior were generated using an annealed Gibbs sampler that sequentially flips individual pixels (and required one million forward model solves). This procedure is computationally expensive and scales exponentially with the number of pixels\footnote{For this reason, the reference posterior was obtained for a discretization of size $64\times64$ and not at the $128\times128$ resolution used by the models compared. Nevertheless, both qualitative structure and quantitative summary statistics suggest that the posterior shown here is a close approximation of the corresponding posterior at higher resolution.}.

\begin{figure}[H]
    \centering

    \begin{subfigure}{\textwidth}
        \centering
    \includegraphics[width=1\linewidth]{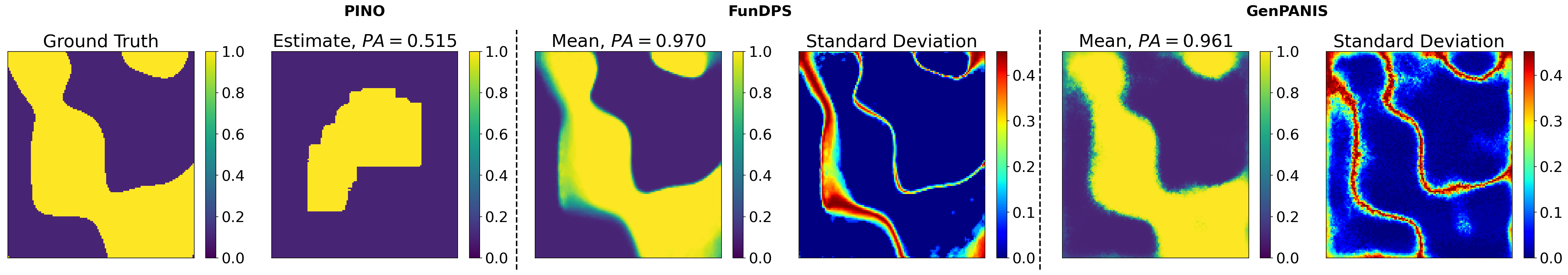}
        \caption{SNR=100}
        \label{fig:snr100_1}
    \end{subfigure}

    \vspace{1em}

    \begin{subfigure}{\textwidth}
        \centering
        \includegraphics[width=1\linewidth]{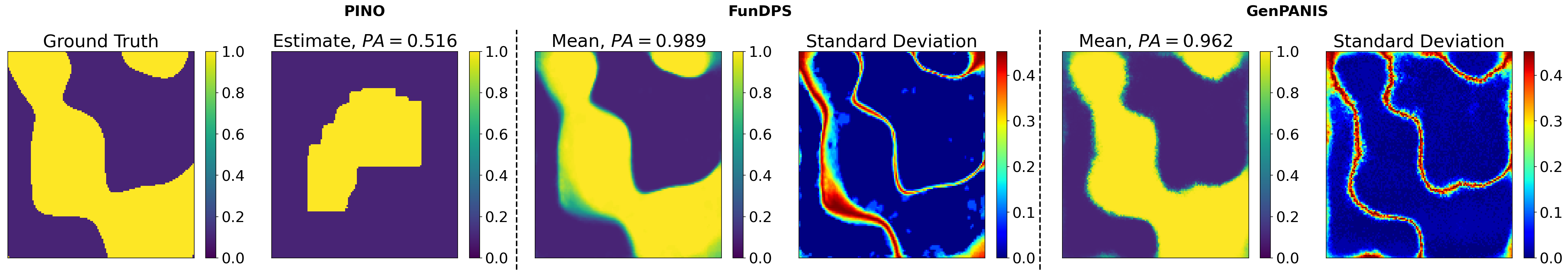}
        \caption{SNR=10}
        \label{fig:snr10}
    \end{subfigure}

    \begin{subfigure}{\textwidth}
        \centering
        \includegraphics[width=1\linewidth]{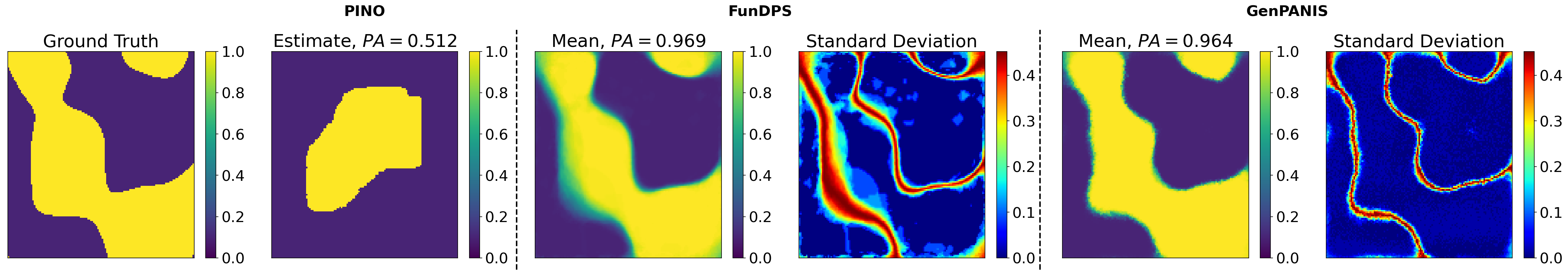}
        \caption{SNR=3}
        \label{fig:snr3}
    \end{subfigure}

    \caption{Solving the Darcy flow inverse problem for various levels of noise by observing the full solution field. No additional residual information is used after training the models.}
    \label{fig:exp1d}
\end{figure}

\begin{figure}[H]
    \centering

    \begin{subfigure}{\textwidth}
        \centering
        \includegraphics[width=1\linewidth]{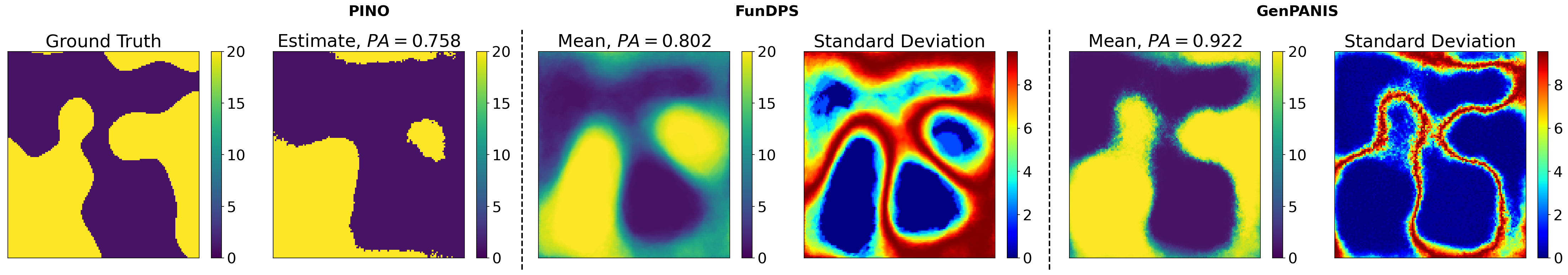}
        \caption{SNR=100}
        \label{fig:snr100_2}
    \end{subfigure}

    \vspace{1em}

    \begin{subfigure}{\textwidth}
        \centering
        \includegraphics[width=1\linewidth]{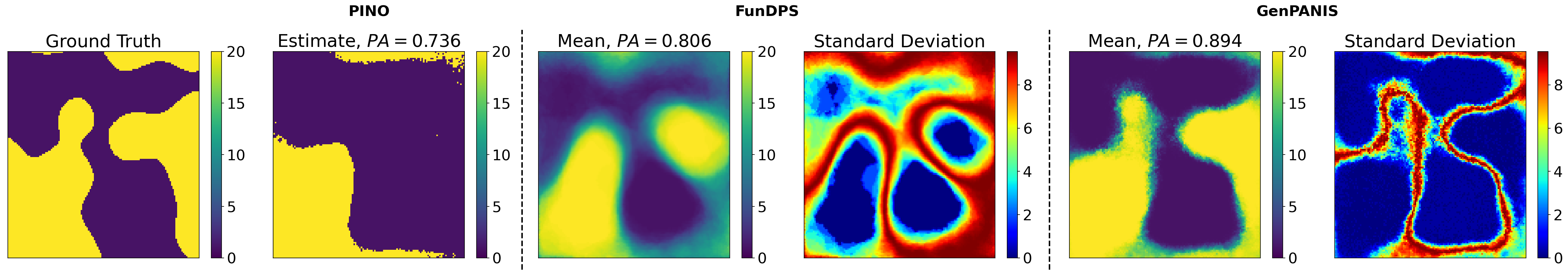}
        \caption{SNR=10}
        \label{fig:snr10}
    \end{subfigure}

    \begin{subfigure}{\textwidth}
        \centering
        \includegraphics[width=1\linewidth]{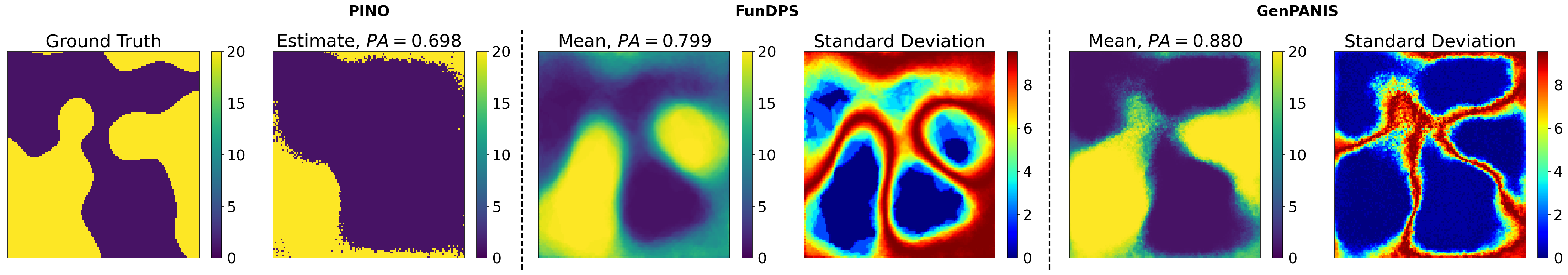}
        \caption{SNR=3}
        \label{fig:snr3}
    \end{subfigure}

    \caption{Solving the Helmholtz equation  inverse problem for various levels of noise by observing the full solution field. No additional residual information is used after training the models.}
    \label{fig:exp1h}
\end{figure}

In Figures \ref{fig:exp1d} and \ref{fig:exp1h}, the predictions of each model are shown for the Darcy flow (\refeq{eq:Darcy}) and  the Helmholtz equation (\refeq{eq:Helmholtz}), respectively, for various levels of noise\footnote{No additional residuals were used for PINO and FunDPS during the prediction phase.}.

In the context of  the Darcy flow  (Figure \ref{fig:exp1d}), we observe that PINO is unable to provide any meaningful prediction, even in the  low-noise case ($SNR=100$).
The diffusion model (FunDPS) provides very accurate mean predictions for any level of noise, but seems to overestimate the posterior  standard deviation. 
By observing the recovered microstructure in the 1st row of Figure \ref{fig:exp1d}, we may visually confirm that GenPANIS is slightly closer to the true posterior from Figure \ref{fig:posteriorExp1} compared to FunDPS. Lastly, GenPANIS performs very competitively with FunDPS across low, medium, and high noise levels (see \ref{appendix:snr}), despite being orders of magnitude lighter in terms of number of parameters and training time.

In the context of the Helmholtz Equation (Figure \ref{fig:exp1h}), we note that PINO provides notably better mean predictions than in the Darcy Flow. Its mean predictions are comparable in accuracy to those of FunDPS, which performs clearly worse in this setting. GenPANIS clearly outperforms all other models on this inverse problem, irrespective of the noise level.

\subsubsection{Use of additional residuals during inference by PINO and FunDPS}
We next investigate whether incorporating additional PDE residual information at prediction time improves inverse performance, albeit at increased computational cost. This experiment clarifies the role of residual-based refinement for models that support it and highlights differences in robustness when observation noise is present.

\begin{figure}[H]
    \centering

    \begin{subfigure}{\textwidth}
        \centering
        \includegraphics[width=1.0\linewidth]{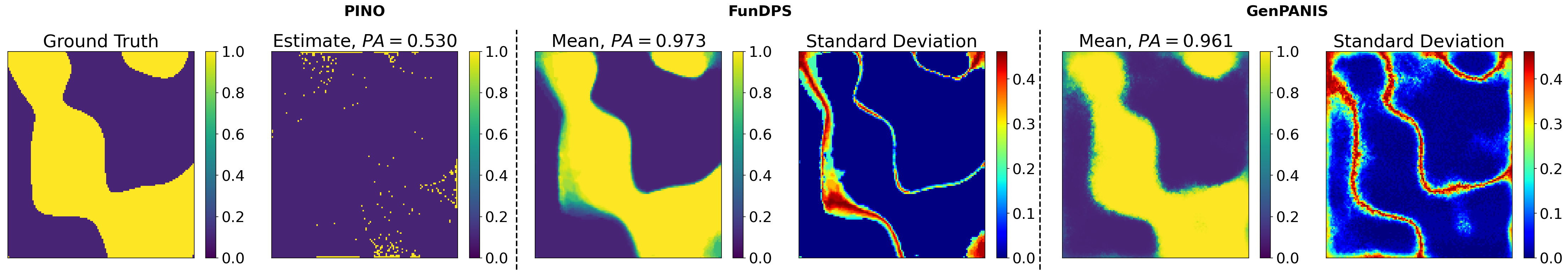}
        \caption{SNR=100}
        \label{fig:snr100f}
    \end{subfigure}

    \vspace{1em}

    \begin{subfigure}{\textwidth}
        \centering
        \includegraphics[width=1.0\linewidth]{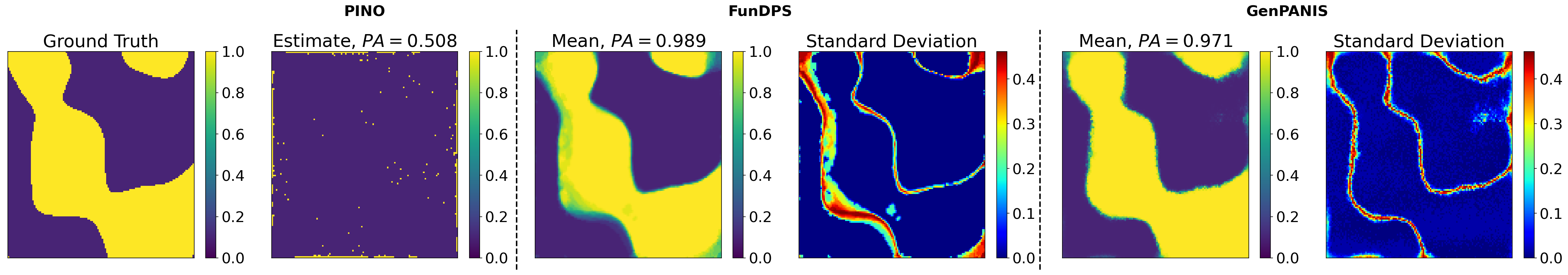}
        \caption{SNR=10}
        \label{fig:snr10f}
    \end{subfigure}

    \begin{subfigure}{\textwidth}
        \centering
        \includegraphics[width=1.0\linewidth]{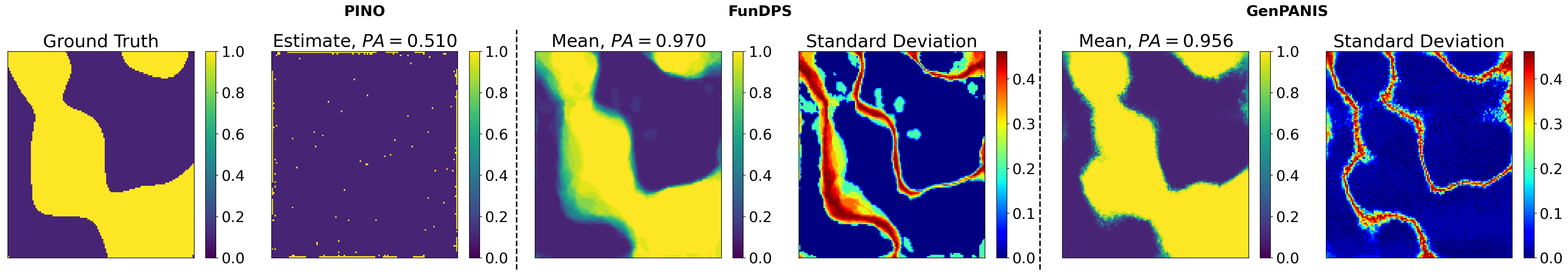}
        \caption{SNR=3}
        \label{fig:snr3f}
    \end{subfigure}

    \caption{Solving the Darcy flow  inverse problem for various levels of noise by observing the full solution field. Additional residuals were used for the PINO and FunDPS models only.}
    \label{fig:exp1df}
\end{figure}

\begin{figure}[H]
    \centering

    \begin{subfigure}{\textwidth}
        \centering
        \includegraphics[width=1.0\linewidth]{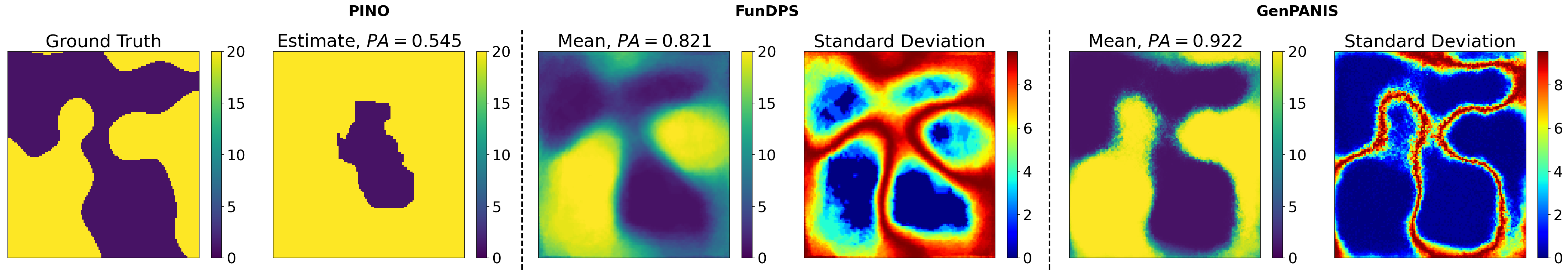}
        \caption{SNR=100}
        \label{fig:snr100f}
    \end{subfigure}

    \vspace{1em}

    \begin{subfigure}{\textwidth}
        \centering
        \includegraphics[width=1.0\linewidth]{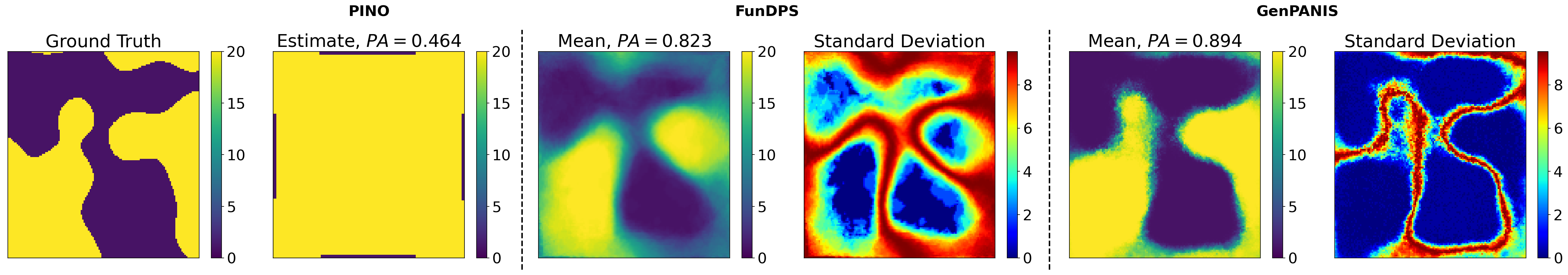}
        \caption{SNR=10}
        \label{fig:snr10f}
    \end{subfigure}

    \begin{subfigure}{\textwidth}
        \centering
        \includegraphics[width=1.0\linewidth]{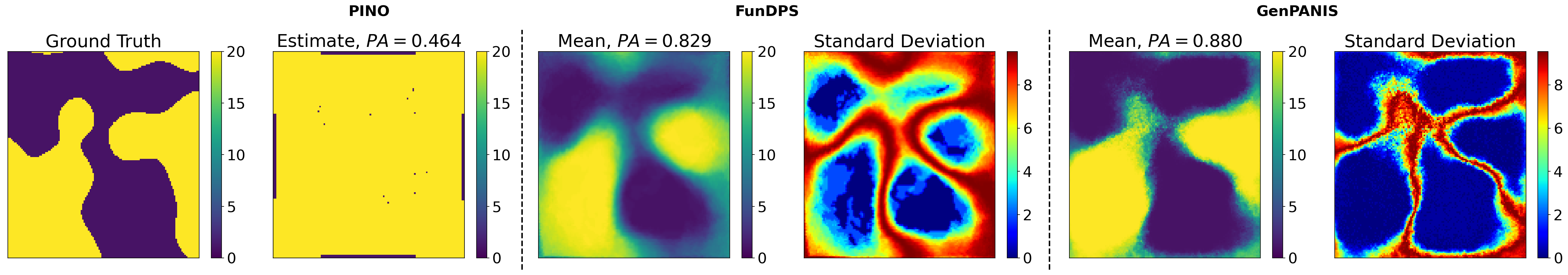}
        \caption{SNR=3}
        \label{fig:snr3f}
    \end{subfigure}

    \caption{Solving the inverse problem for various levels of noise by observing the full solution field. Additional residuals were used for the PINO and FunDPS models only.}
    \label{fig:exp1hf}
\end{figure}

In Figures \ref{fig:exp1df} and \ref{fig:exp1hf}, we provide the same results as before, but in the case where additional residuals were used during the inference phase. PINO completely collapses from the moment any non-negligible noise is introduced in the observations. Furthermore, when additional residuals are employed for FunDPS, and especially in the Helmholtz problem, the posterior estimates are improved by a significant margin, but not enough to surpass the performance of GenPANIS, even though the latter does not employ any additional residuals during inference. The additional residuals did not significantly impact the performance of FunDPS for the Darcy Flow equation.

\paragraph{Partial Observation on a Fixed Grid}

We then consider inverse problems with spatially sparse measurements taken on fixed regular grids of decreasing resolution. These experiments assess each model’s ability to interpolate missing information and recover the input field when no information is available over large portions of the problem domain.

\begin{figure}[H]
    \centering

    \begin{subfigure}{\textwidth}
        \centering
        \includegraphics[width=1.0\linewidth]{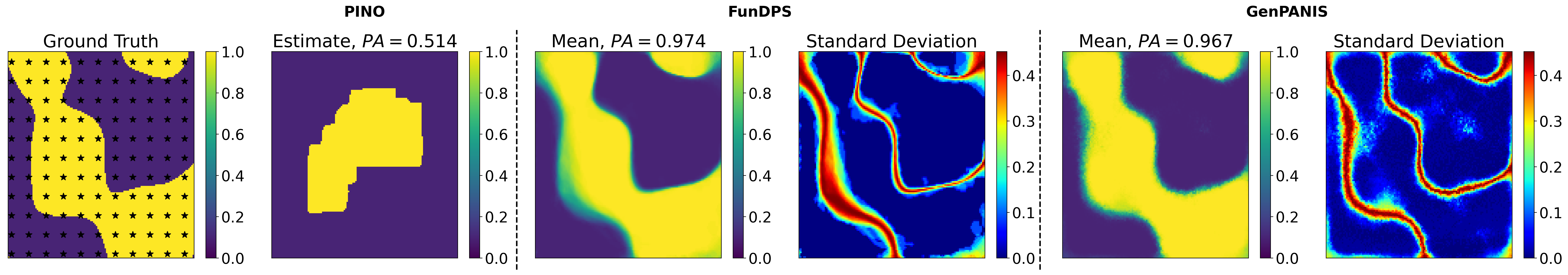}
        \caption{Fixed Grid: 11x11}
        \label{fig:snr100_11x11}
    \end{subfigure}

    \vspace{1em}

    \begin{subfigure}{\textwidth}
        \centering
        \includegraphics[width=1.0\linewidth]{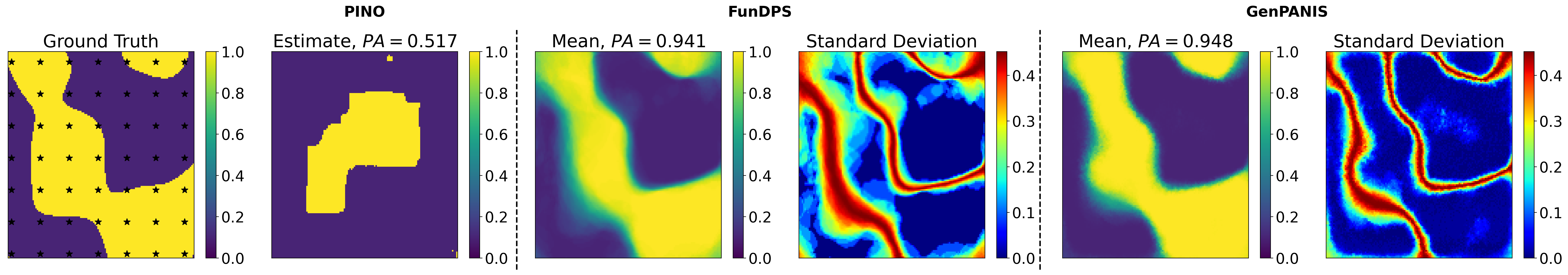}
        \caption{Fixed Grid: 7x7}
        \label{fig:snr100_7x7}
    \end{subfigure}

    \vspace{1em}

    \begin{subfigure}{\textwidth}
        \centering
        \includegraphics[width=1.0\linewidth]{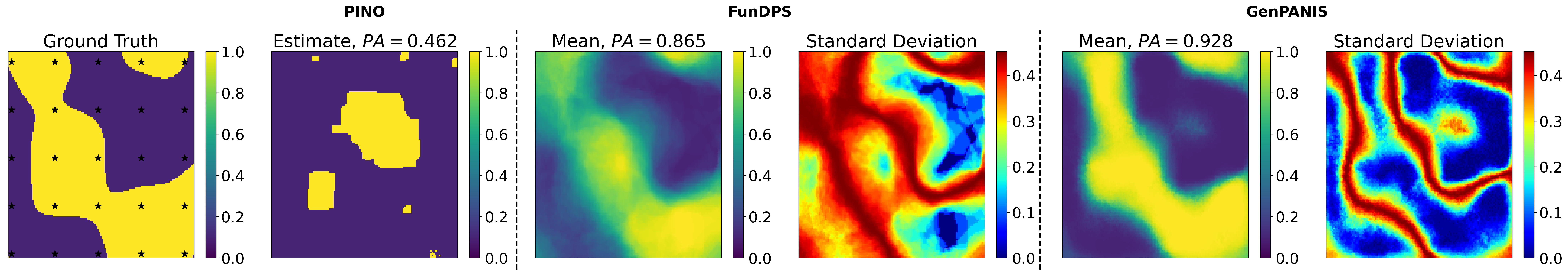}
        \caption{Fixed Grid: 5x5}
        \label{fig:snr100_5x5}
    \end{subfigure}

    \caption{Solving the Darcy flow inverse problem for various observation grids and $SNR=100$. No additional residual information is used after training the models.}
    \label{fig:exp2d}
\end{figure}

\begin{figure}[H]
    \centering

    \begin{subfigure}{\textwidth}
        \centering
        \includegraphics[width=1.0\linewidth]{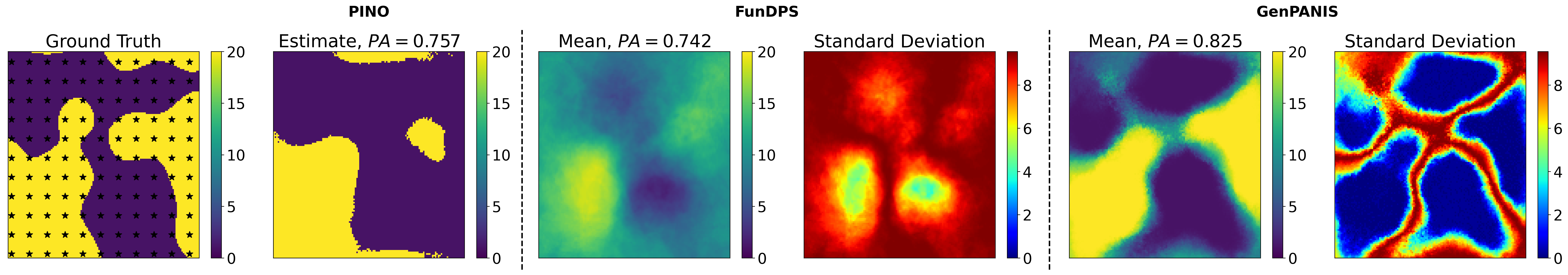}
        \caption{Fixed Grid: 11x11}
        \label{fig:snr100_11x11}
    \end{subfigure}

    \vspace{1em}

    \begin{subfigure}{\textwidth}
        \centering
        \includegraphics[width=1.0\linewidth]{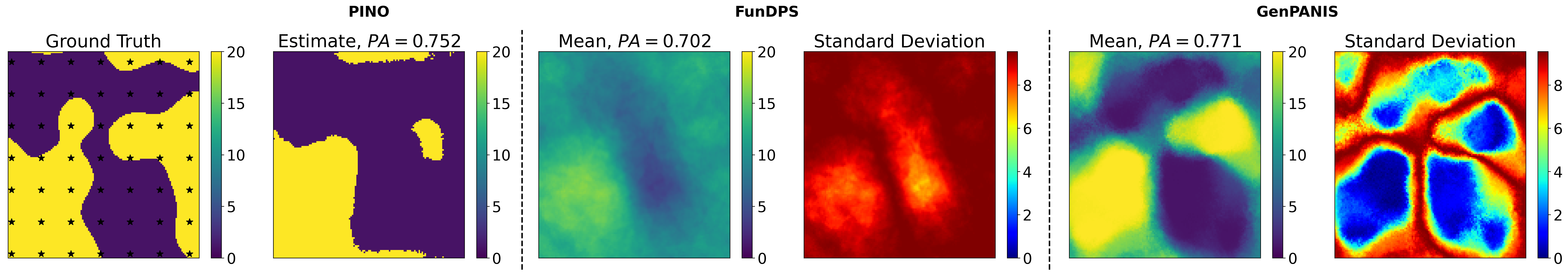}
        \caption{Fixed Grid: 7x7}
        \label{fig:snr100_7x7}
    \end{subfigure}

    \vspace{1em}

    \begin{subfigure}{\textwidth}
        \centering
        \includegraphics[width=1.0\linewidth]{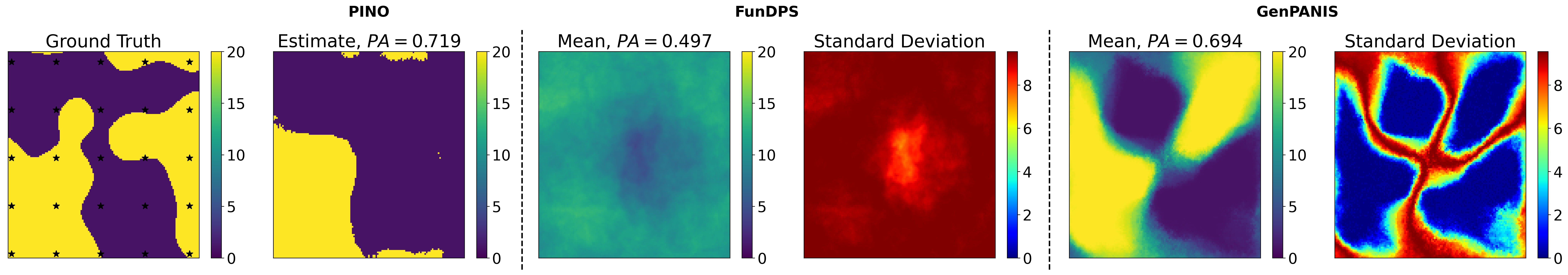}
        \caption{Fixed Grid: 5x5}
        \label{fig:snr100_5x5}
    \end{subfigure}

    \caption{Solving the Helmholtz equation inverse problem for various  observation grids and $SNR=100$. No additional residual information is used after training the models.}
    \label{fig:exp2h}
\end{figure}

In Figures \ref{fig:exp2d} and \ref{fig:exp2h}, we fix the noise level to low ($SNR=100$), and we investigate the performance of each model when observations are available on regular grids of 11$\times$11, 7$\times$7, and 5$\times$5, respectively\footnote{We note that in  these and all the subsequent illustrations, no additional residuals are used during the prediction phase.}. 
As expected, the quality of the predictions deteriorates across all models as observations become sparser. PINO fails completely to recover the input microstructure for the Darcy problem, and it performs poorly for the Helmholtz case. The diffusion model FunDPS performs well for the Darcy flow as long as the observations are not too sparse. However, its performance on the Helmholtz equation is even worse than PINO's  in all cases examined. For this experiment, GenPANIS is clearly better, as it maintains its mean accuracy across all cases (see e.g. PA scores) and produces increased posterior uncertainty estimates with reduced observables, as one would expect.

\paragraph{Randomly Selected Partial Observations}
Finally, we consider randomly distributed observation locations. Compared to fixed grids, this setting introduces regions with little or no local information, thereby posing a more challenging and realistic inverse problem.

\begin{figure}[H]
    \centering

    \begin{subfigure}{\textwidth}
        \centering
        \includegraphics[width=1.0\linewidth]{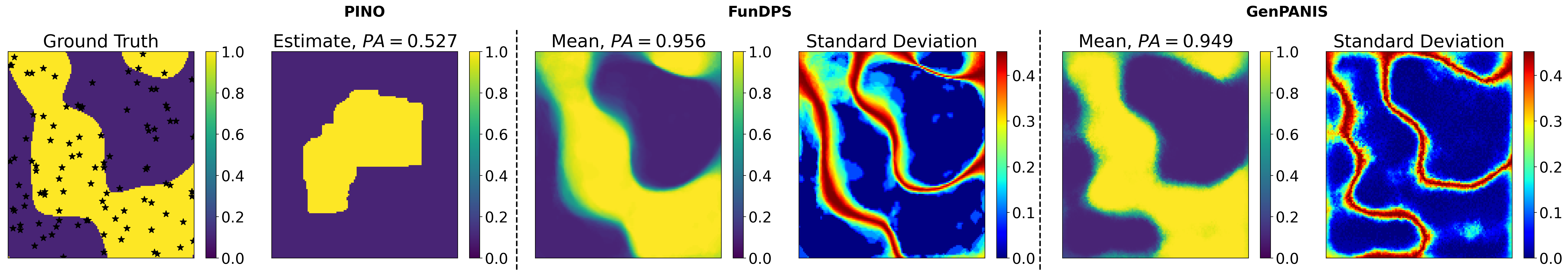}
        \caption{100 random observations}
        \label{fig:snr100_r100}
    \end{subfigure}

    \vspace{1em}

    \begin{subfigure}{\textwidth}
        \centering
        \includegraphics[width=1.0\linewidth]{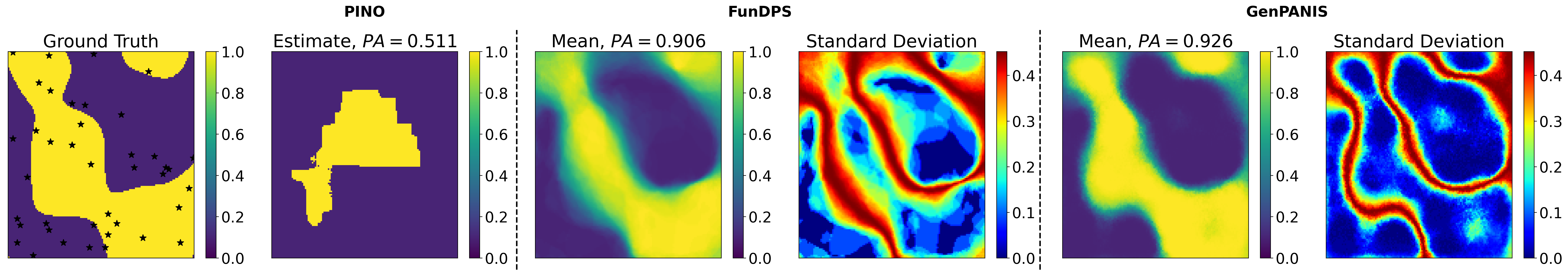}
        \caption{40 random observations}
        \label{fig:snr100_r40}
    \end{subfigure}

    \vspace{1em}

    \begin{subfigure}{\textwidth}
        \centering
        \includegraphics[width=1.0\linewidth]{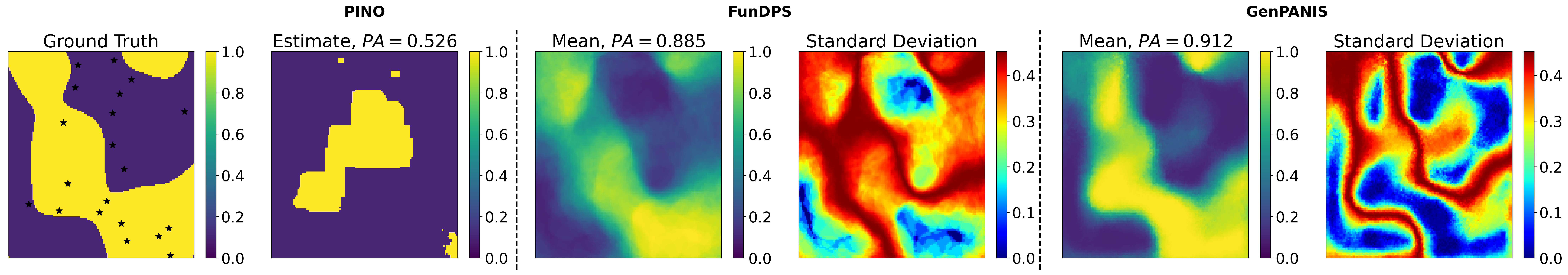}
        \caption{20 random observations}
        \label{fig:snr100_r20}
    \end{subfigure}

    \caption{Solving the Darcy flow inverse problem for various randomly selected observation points and $SNR=100$. No additional residual information is used after training the models.}
    \label{fig:exp3d}
\end{figure}

\begin{figure}[H]
    \centering

    \begin{subfigure}{\textwidth}
        \centering
        \includegraphics[width=1.0\linewidth]{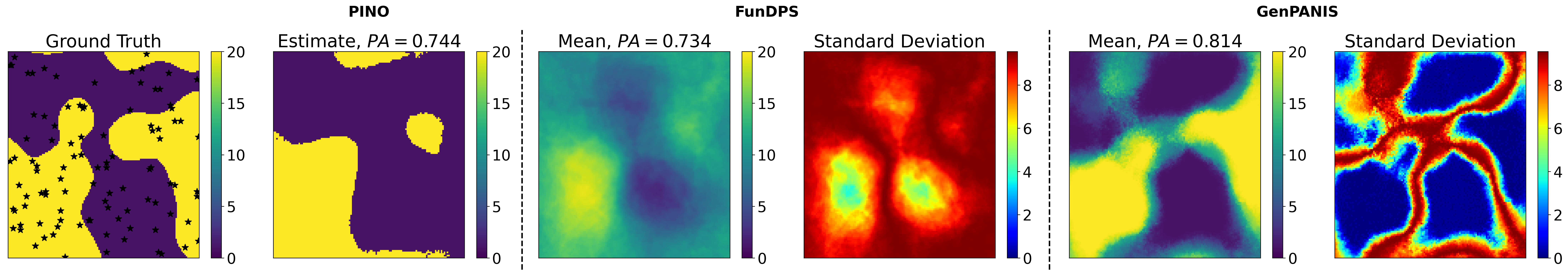}
        \caption{100 random observations}
        \label{fig:snr100_r100}
    \end{subfigure}

    \vspace{1em}

    \begin{subfigure}{\textwidth}
        \centering
        \includegraphics[width=1.0\linewidth]{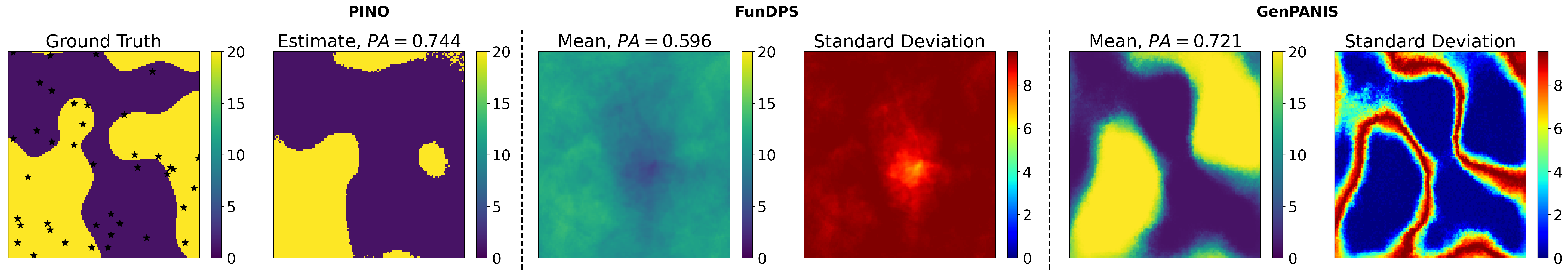}
        \caption{40 random observations}
        \label{fig:snr100_r40}
    \end{subfigure}

    \vspace{1em}

    \begin{subfigure}{\textwidth}
        \centering
        \includegraphics[width=1.0\linewidth]{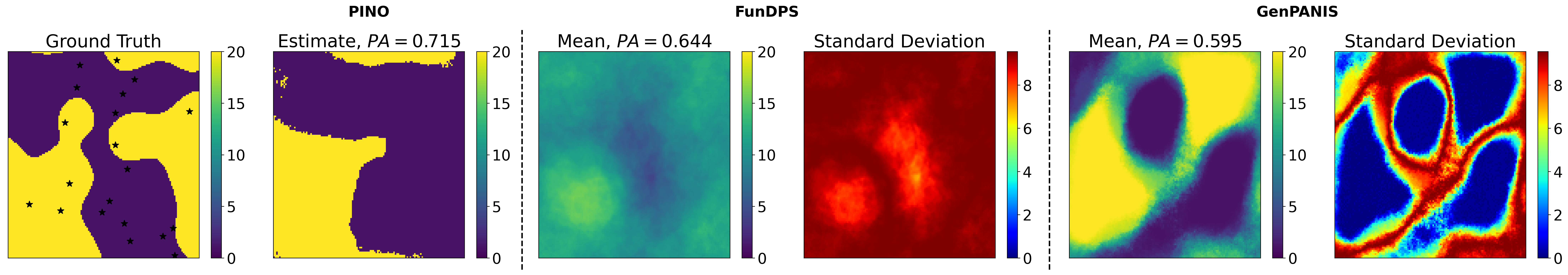}
        \caption{20 random observations}
        \label{fig:snr100_r20}
    \end{subfigure}

    \caption{Solving the Helmholtz equation inverse problem for various randomly selected observation points and $SNR=100$. No additional residual information is used after training the models.}
    \label{fig:exp3h}
\end{figure}

For the illustrations in the current subsection, the noise level is set to $SNR=100$, and we test all models on the same inverse problem using 100, 40, and 20 randomly sampled observations of the output, respectively. The conclusions drawn from these experiments are similar to those of the previous subsection. In this very challenging test, the only model that predicts with decent accuracy is GenPANIS, as shown in Figures \ref{fig:exp3d} and \ref{fig:exp3h}.

\paragraph{Effect of Number of Labeled Training Data}
\label{sec:exp7}

\begin{figure}[H]
    \centering
    \includegraphics[width=1.0\linewidth]{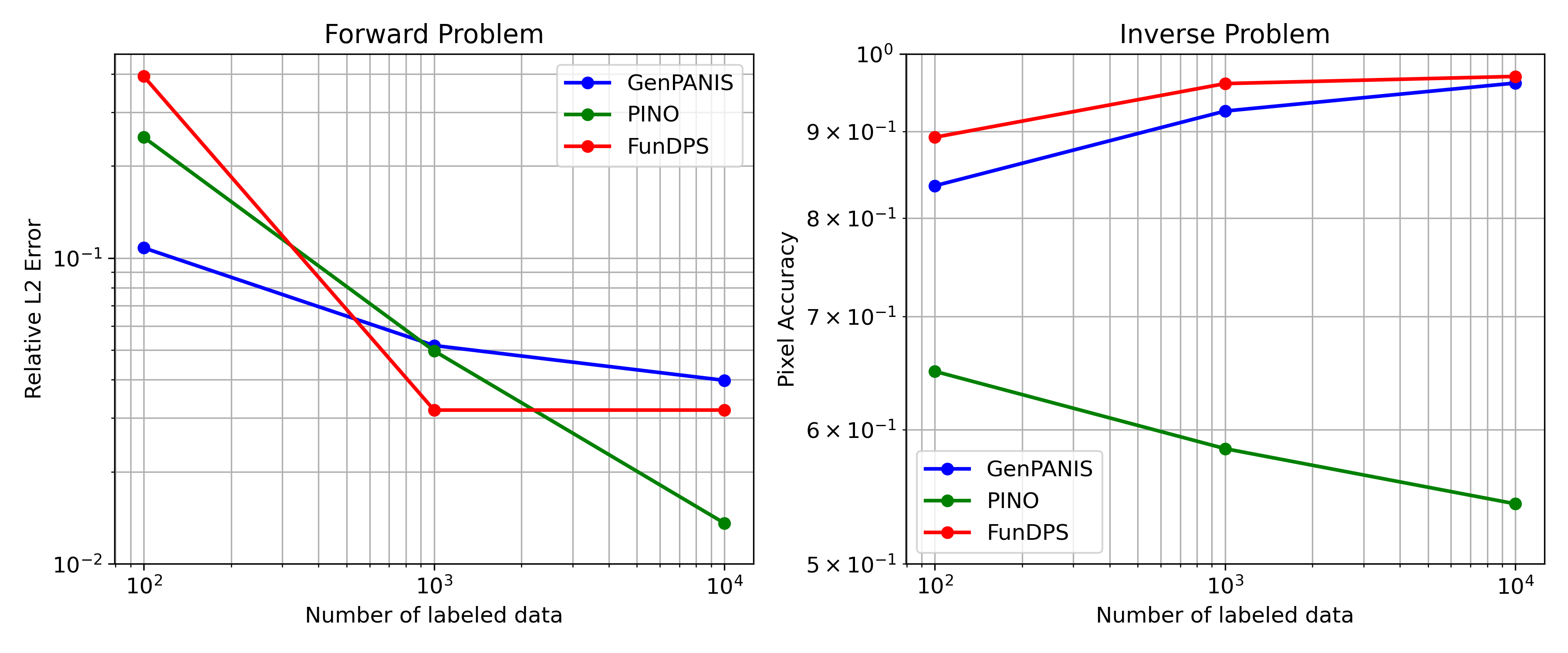}
    \caption{Performance of the examined models, in terms of $L_2$ error $\epsilon$ (forward problem, same instance as Figure \ref{fig:exp1d}, noiseless) and pixelwise accuracy $PA$ (inverse problem, same instance as Figure \ref{fig:exp1d}, $SNR=100$), when different number of labeled data is employed during training.}
    \label{fig:effOfLab}
\end{figure}

We next study how each model's predictive estimates are affected by the  number of labeled training data (i.e. PDE input/output pairs). We investigate their performance both in forward and inverse problems (Figure \ref{fig:effOfLab}). In the former case, GenPANIS saturates the fastest, but PINO  achieves the highest predictive accuracy. In the context of inverse problems, however, PINO seems to struggle, whereas FunDPS and GenPANIS achieve comparably high PA-scores and seem to saturate fairly quickly.

\subsubsection{Out-of-Distribution Inverse Problems}
We now turn to inverse problems that depart from the training distribution. These experiments test the robustness and generalization capabilities of the models when faced with conditions not encountered during training.

\paragraph{Out-of-Distribution Input-Output Pairs}

\begin{figure}[H]
    \centering
    \includegraphics[width=1.0\linewidth]{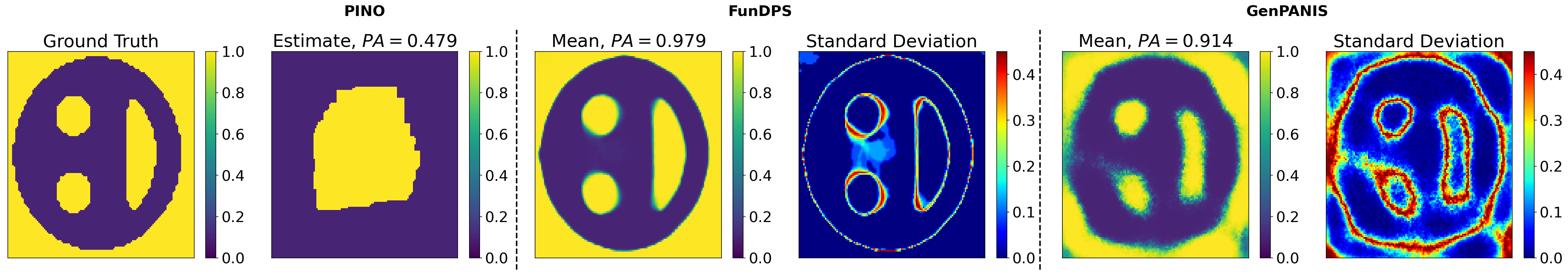}
    \caption{Solving the Darcy flow inverse problem for an out-of-distribution input-output pair. The solution field is noisy ($SNR=100$) and fully observed. No additional residual information is used after training the models.}
    \label{fig:exp4}
\end{figure}

We first evaluate performance on inverse problems where the ground truth  is far removed from  the training data. During the experiment shown in Figure \ref{fig:exp4} for the Darcy flow, we examine the performance of each model in solving an inverse problem with full, low-noise observations, when the input microstructure is clearly outside the training dataset.  We observe that PINO completely fails to recover the underlying microstructure, whereas FunDPS and GenPANIS demonstrate satisfactory performance. Nevertheless, FunDPS appears to offer a higher-quality solution to the inverse problem, both in terms of mean accuracy and uncertainty bounds.

\paragraph{Out-of-Distribution Boundary Conditions}
We then examine inverse problems generated using boundary conditions that differ from those seen during training. This setting tests whether the models can adapt to new physical constraints without retraining.

In particular, we consider the Darcy flow and Helmholtz problems with the following boundary conditions: 

\begin{equation}
    u_0(s_1,s_2)=
    \begin{cases}
    \displaystyle 5(s_1 + s_2), 
    & \text{if } s_1 = 0,\; 0 < s_2 < 1, \\
    
    \displaystyle 5(s_1 + s_2),
    & \text{if } s_2 = 1,\; 0 < s_1 < 1, \\
    
    \displaystyle 5(s_1 + s_2),
    & \text{if } s_1 = 1,\; 0 < s_2 < 1, \\
    
    \displaystyle 5(s_1 + s_2),
    & \text{if } s_2 = 0,\; 0 < s_1 < 1.
    \end{cases}
    \label{eq:outBc}
\end{equation}
which differ from those in \refeq{eq:Darcy} and \refeq{eq:Helmholtz}, which were used to generate the training data.

\begin{figure}[!t]
    \centering
    \includegraphics[width=1.0\linewidth]{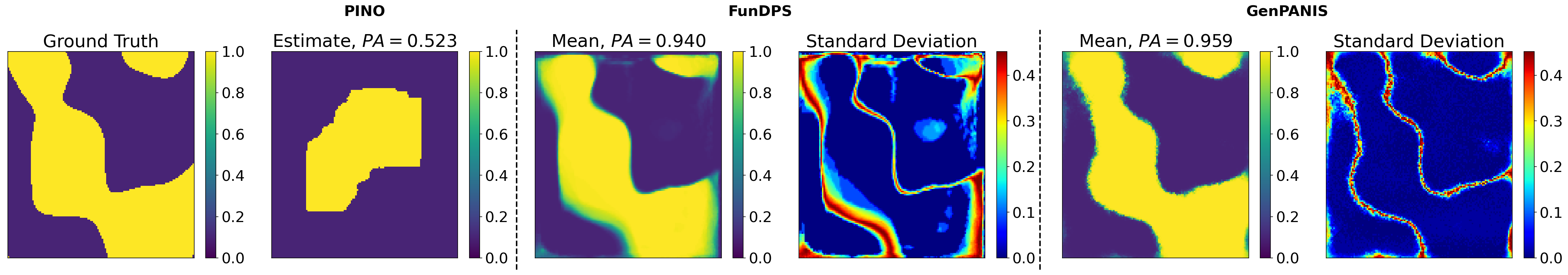}
    \caption{Solving the Darcy flow inverse problem for out-of-distribution Boundary conditions described in Equation \ref{eq:outBc}. The solution field is noisy ($SNR=100$) and fully observed. No additional residual information is used after training the models.}
    \label{fig:exp5d}
\end{figure}

\begin{figure}[!t]
    \centering
    \includegraphics[width=1.0\linewidth]{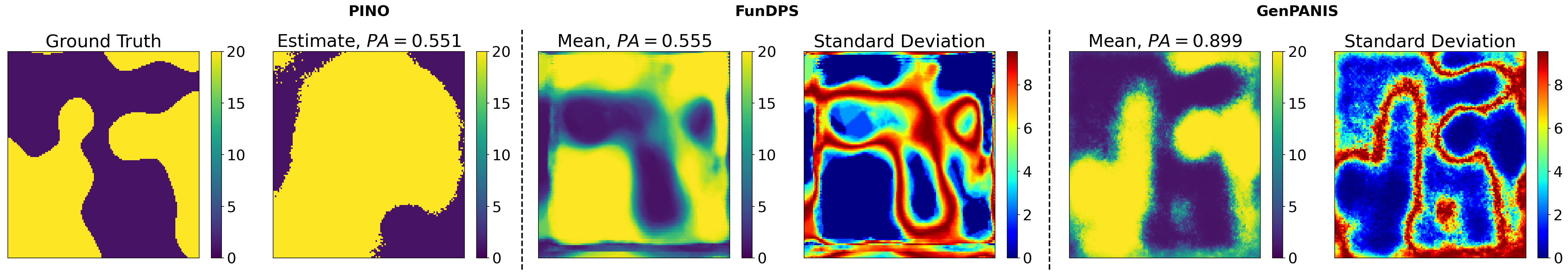}
    \caption{Solving the Helmholtz equation inverse problem for out-of-distribution Boundary conditions described in Equation \ref{eq:outBc}. The solution field is noisy ($SNR=100$) and fully observed. No additional residual information is used after training the models.}
    \label{fig:exp5h}
\end{figure}

\noindent In Figures \ref{fig:exp5d} and \ref{fig:exp5h}, we observe that PINO fails to correctly identify the ground truth in both cases. FunDPS demonstrates good performance for the Darcy Flow equation, but very poor performance for the Helmholtz equation. In contrast, GenPANIS retains its accuracy and produces estimates as accurate as in the  in-distribution tests. This can be attributed to the embedded coarse-grained model, which can automatically adapt to different boundary conditions. 

\begin{figure}[H]
    \centering
\includegraphics[width=1.0\linewidth]{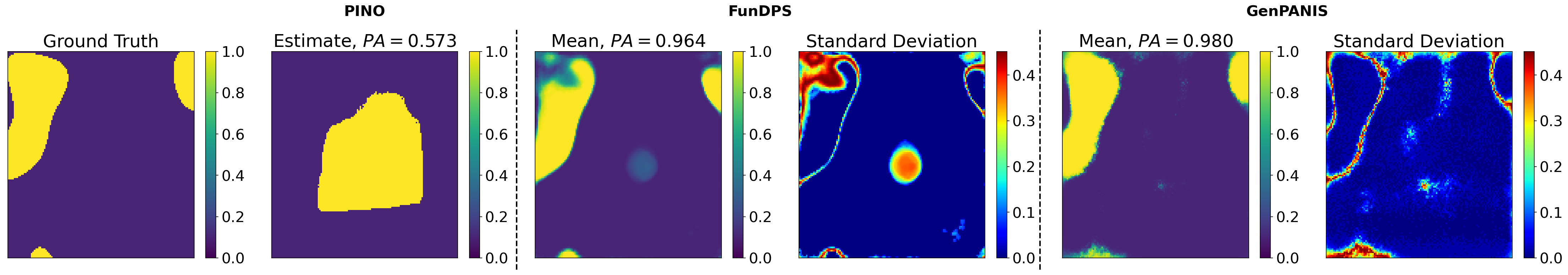}
    \caption{Solving the Darcy flow inverse problem for out-of-distribution volume fraction (VF=10\%). The solution field is noisy ($SNR=100$) and fully observed. No additional residual information is used after training the models.}
    \label{fig:exp6d}
\end{figure}

\begin{figure}[H]
    \centering
   \includegraphics[width=1.0\linewidth]{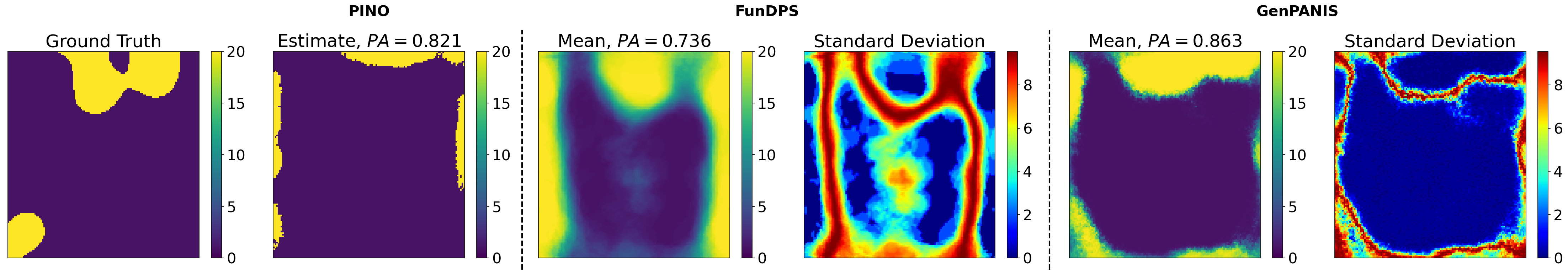}
    \caption{Solving the Helmholtz equation inverse problem for out-of-distribution volume fraction (VF=10\%). The solution field is noisy ($SNR=100$) and fully observed. No additional residual information is used after training the models.}
    \label{fig:exp6h}
\end{figure}

\paragraph{Out-of-Distribution Volume Fractions}
Next, we assess generalization across microstructures with volume fractions that deviate from those used during training. In particular, the training data employed microstructures with a volume fraction of $50\%$, whereas the following experiments used ground-truth microstructures with a volume fraction of $10\%$. For both the Darcy flow (Figure \ref{fig:exp6d}) and the Helmholtz equation (Figure \ref{fig:exp6h}), we see that GenPANIS exhibits superior accuracy in both the posterior mean and the posterior variance.

\subsubsection{Effect of using different mixtures of data}
\label{sec:exp8}
In the final set of experiments, we investigate the capability of GenPANIS to employ less expensive types of training data, namely, unlabeled and virtual (see section \ref{sec:dataTypes}), in addition to or in lieu of the labeled data employed thus far.

\paragraph{The Effect of using Unlabeled Data}
We investigate the impact of augmenting a small labeled dataset ($N_l=100$) with additional unlabeled samples and quantify the resulting gains in both forward and inverse performance.

\begin{figure}[H]
    \centering

    \begin{subfigure}{\textwidth}
        \centering
        \includegraphics[width=1.0\linewidth]{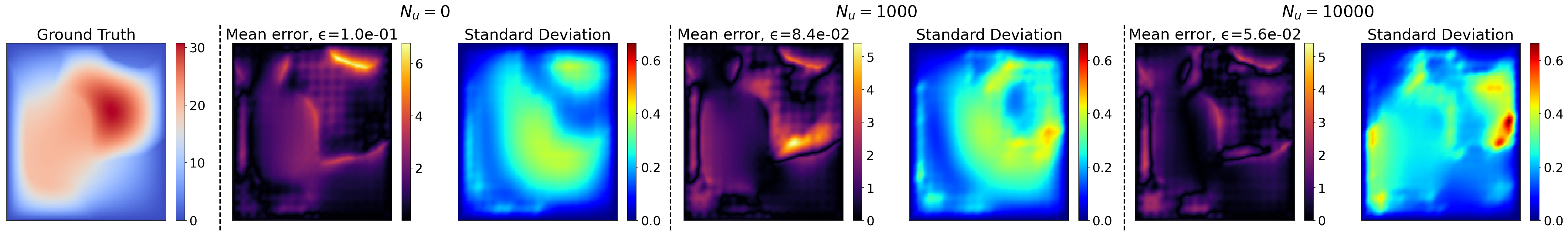}
        \caption{Forward Problem}
        \label{fig:effOfUnForward}
    \end{subfigure}

    \vspace{1em}

    \begin{subfigure}{\textwidth}
        \centering
        \includegraphics[width=1.0\linewidth]{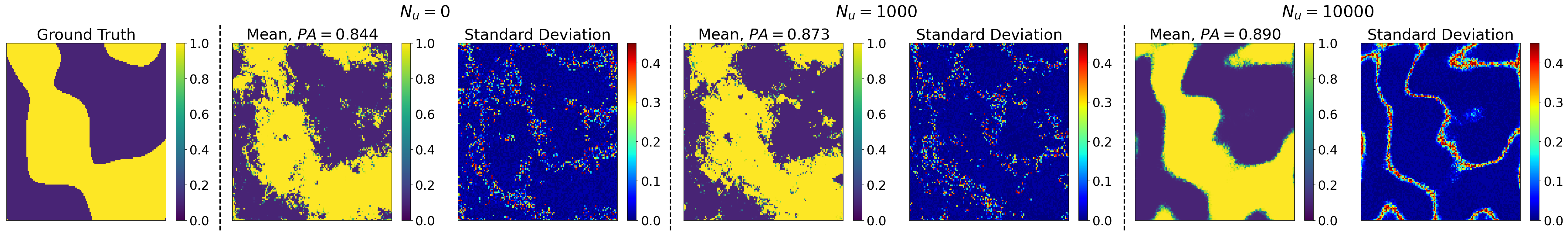}
        \caption{Inverse Problem}
        \label{fig:effOfUnInverse}
    \end{subfigure}
    \caption{Effect of using unlabeled data on top of a small labeled dataset ($N_l=100$). Full observations were used for both the forward and inverse problems, while for the latter, the observation noise was equal to $SNR=100$.}
    \label{fig:exp7a}
\end{figure}

\begin{figure}[H]
    \centering
    \includegraphics[width=0.7\linewidth]{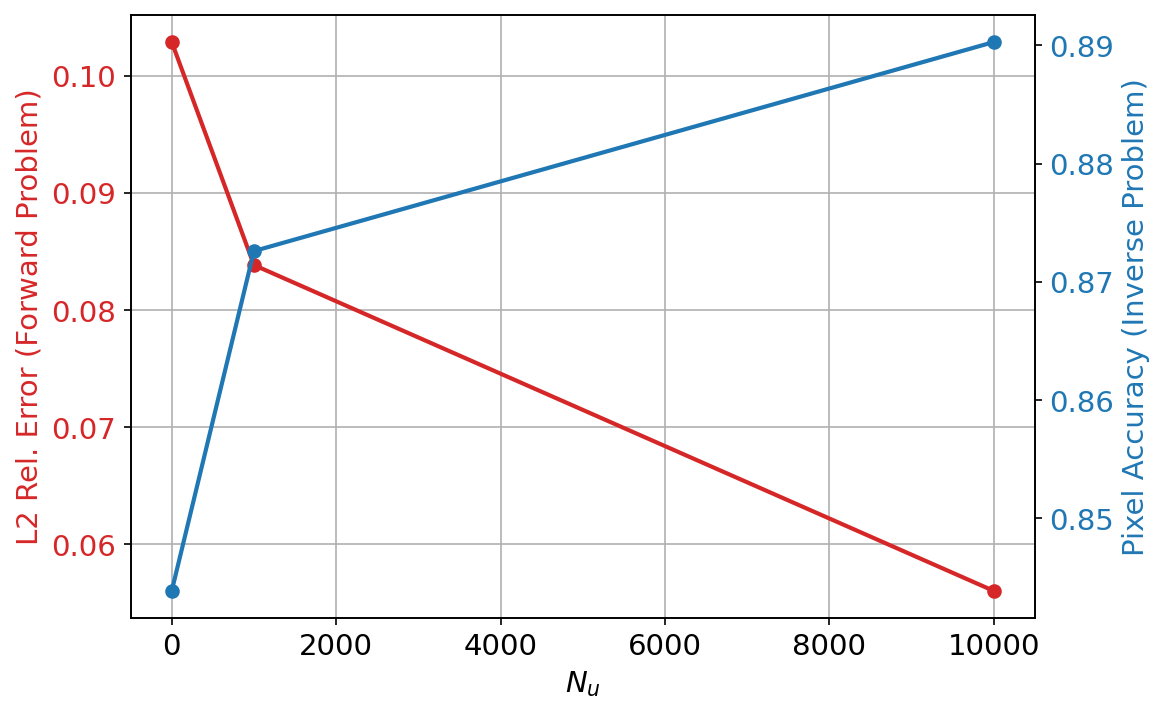}

    \caption{Predictive performance of the model, as more  unlabeled training data are added to $N_l=100$ labeled training data.}
    \label{fig:exp7b}
\end{figure}

We note that neither PINO nor FunDPS can incorporate unlabeled data, and thus, a direct comparison with GenPANIS is not possible. In Figure \ref{fig:exp7a}, we can  observe the massive improvement that the gradual incorporation of $1000$ and $10,000$ unlabeled data can have on the performance for both the forward \ref{fig:effOfUnForward} and the inverse \ref{fig:effOfUnInverse} problem. In particular, and as it is summarized in Figure \ref{fig:exp7b}, employing $1000$ unlabeled data decreases the $L_2$ error  by $ 18.5 \%$ for the forward problem and increases the pixel accuracy for the inverse problem by $ 3.4\%$. The improvements are even more striking when using $10,000$ unlabeled data, with the respective percentages being $45.6\%$ and $5.5\%$. These gains in accuracy are even more important if we consider that using unlabeled data involves practically no additional pre-processing or training cost.

\paragraph{Effect of Combining Labeled Data with Virtual Observables}
Lastly, we study the use of virtual observables derived from the governing equations and analyze how combining labeled data with physics-based information influences model accuracy and data efficiency.

We first note that FunDPS and most other diffusion models \cite{huang2024diffusionpde} lack the ability to train using information directly from the PDE, in the form of residuals. PINO can employ strong residuals during training, but in an inflexible manner that depends on the labeled dataset; i.e., residuals are computed only for input field instances present in the labeled dataset used during training.

We report the performance of GenPANIS for  forward and inverse problems in Tables \ref{tab:labeled-VO-f} and \ref{tab:labeled-VO-i}, respectively, for different combinations of labeled $N_l$ and virtual $N_v$ training data. Firstly, we observe that the addition of any of these two data types improves the model's performance. Furthermore, we observe that the addition of labeled data  yields slightly better performance than the addition of an equal number of virtual data. Nevertheless, the slightly inferior benefit of virtual observables is compensated for by the fact that no forward simulations are needed in order to generate them, i.e., the forward problem never needs to be solved.

\begin{table}[!t]
\centering
\begin{tabular}{|c|c|c|c|c|}
\hline
$N_l$/$N_v$ & $N_v=0$ & $N_v=100$ & $N_v=1000$ & $N_v=10000$ \\ \hline
$N_l=0$ & - & $1.547 \cdot 10^{-1}$ & $1.189 \cdot 10^{-1}$ & $0.963 \cdot 10^{-1}$ \\ \hline
$N_l=100$ & $1.029 \cdot 10^{-1}$ & $0.983 \cdot 10^{-1}$ & $0.967 \cdot 10^{-1}$ & $0.925 \cdot 10^{-1}$  \\ \hline
$N_l=1000$ & $0.839 \cdot 10^{-1}$ & $0.684 \cdot 10^{-1}$ & $0.638 \cdot 10^{-1}$ & $0.553 \cdot 10^{-1}$ \\ \hline
$N_l=10000$ & $0.560 \cdot 10^{-1}$ & $0.428 \cdot 10^{-1}$ & $0.394 \cdot 10^{-1}$ & $0.387 \cdot 10^{-1}$ \\ \hline
\end{tabular}
\caption{$L_2$ error $\epsilon$ for solving the forward problem  (same instance as \ref{fig:exp1d}) for different combinations of labeled ($N_l$) and virtual ($N_v$) data.}
\label{tab:labeled-VO-f}
\end{table}

\begin{table}[!t]
\centering
\begin{tabular}{|c|c|c|c|c|}
\hline
$N_l$|$N_v$ & $N_v=0$ & $N_v=100$ & $N_v=1000$ & $N_v=10000$ \\ \hline
$N_l=0$ & - & $0.813$ & $0.849$ & $0.854$ \\ \hline
$N_l=100$ & $0.844$ & $0.853$ & $0.859$ & $0.883$  \\ \hline
$N_l=1000$ & $0.873$ & $0.906$ & $0.918$ & $0.922$ \\ \hline
$N_l=10000$ & $0.890$ & $0.955$ & $0.966$ & $0.966$ \\ \hline
\end{tabular}
\caption{Pixelwise Accuracy for solving the inverse problem of (same instance as \ref{fig:exp1d}, $SNR=100$) for different combinations of labeled ($N_l$) and virtual ($N_v$) data.}
\label{tab:labeled-VO-i}
\end{table}

\section{Conclusions}
\label{sec:conclusions}

We have introduced GenPANIS, a unified physics-aware probabilistic generative model for solving forward and inverse problems in multiphase media with discrete material fields. 
The framework addresses fundamental challenges in multiphase media by enabling gradient-based inference on discrete microstructures without continuous relaxation, capabilities essential for both inverse analysis and inverse design applications.

The numerical experiments presented in Section \ref{sec:numerical} demonstrate several key advantages of GenPANIS over state-of-the-art methods including PINO \cite{li2021physics} and FunDPS \cite{yao2025guided}:

\begin{itemize}
    \item \textbf{Superior performance on challenging inverse problems}: GenPANIS consistently outperforms competing methods on inverse problems with sparse observations (Figures \ref{fig:exp2d}, \ref{fig:exp2h}, \ref{fig:exp3d}, \ref{fig:exp3h}) and high noise levels (Figures \ref{fig:exp1d}, \ref{fig:exp1h}), particularly for the Darcy equation where it maintains high pixel accuracy even under extreme conditions (20 random observations, PA $> 0.9$). Unlike PINO, which fails catastrophically in the presence of noise, and FunDPS, which struggles with spatially sparse measurements, GenPANIS provides accurate posterior mean estimates and well-calibrated uncertainty quantification across all tested scenarios.
    
    \item \textbf{Superior out-of-distribution generalization}: The embedded physics-aware architecture enables GenPANIS to extrapolate reliably to conditions not encountered during training. Most notably, when boundary conditions differ substantially from those in training (Figures \ref{fig:exp5d}, \ref{fig:exp5h}), GenPANIS maintains accuracy, while PINO fails completely and FunDPS deteriorates significantly. Similar robustness is observed for out-of-distribution volume fractions (Figures \ref{fig:exp6d}, \ref{fig:exp6h}) and microstructure configurations (Figure \ref{fig:exp4}). This capability is critical for inverse design applications, where iterative optimization algorithms \cite{aage2017giga, sosnovik2019neural} inevitably generate intermediate states far from the training distribution.
    
    \item \textbf{Significant computational efficiency}: GenPANIS achieves superior inverse problem performance with $\sim 5$ times  fewer parameters than PINO and $\sim 63$ times fewer than FunDPS (Table \ref{tab:modelStats}), resulting in training times that are $\sim 64$ and $\sim 74$ times faster, respectively (0.6 hours vs. 38-45 hours for 10,000 labeled samples). This efficiency stems from the lightweight decoder design combined with the informative normalizing flow prior, which concentrates model capacity in the latent representation rather than in high-dimensional decoders. The model saturates quickly even with limited latent dimensions ($d_z \geq 30$, Table \ref{tab:paramDimz}), demonstrating the effectiveness of the physics-informed bottleneck.
    
    \item \textbf{Flexible data utilization without ad-hoc weighting}: The unified probabilistic formulation (Section \ref{sec:dataTypes}) accommodates unlabeled microstructures, virtual observations from physics residuals, and labeled pairs through a rigorous evidence lower bound that automatically balances contributions from each data type. As demonstrated in Section \ref{sec:exp8}, augmenting 100 labeled samples with 10,000 unlabeled microstructures reduces forward prediction error by 45.6\% and improves inverse pixel accuracy by 5.5\% (Figure \ref{fig:exp7b}), with negligible additional computational cost. Similarly, virtual observables derived purely from PDE residuals—requiring no forward solves—substantially improve performance when combined with labeled data (Tables \ref{tab:labeled-VO-f}, \ref{tab:labeled-VO-i}). This flexibility is particularly valuable for design applications where exhaustive forward simulations are prohibitive \cite{bostanabad2018computational}.
    
    \item \textbf{Unified bidirectional framework}: Both forward prediction and inverse inference operate within the same trained model through conditioning in latent space (Section \ref{sec:prediction}), without retraining or architectural modification. While forward prediction accuracy is comparable to specialized methods (Figure \ref{fig:exp1_forward}), the true strength lies in inverse problems where GenPANIS consistently outperforms dedicated inverse solvers. This bidirectionality is essential for iterative design workflows requiring repeated forward evaluation and inverse synthesis.
    
    \item \textbf{Principled uncertainty quantification}: Unlike deterministic methods or models relying on heuristic guidance \cite{xu_rethinking_2025}, GenPANIS provides well-calibrated posterior distributions over discrete microstructures. Comparison against reference posteriors obtained via expensive MCMC (Figure \ref{fig:posteriorExp1}) confirms that GenPANIS captures posterior structure accurately while requiring only a few gradient-based sampling steps in low-dimensional latent space. Posterior variance estimates appropriately increase with observation sparsity (Figures \ref{fig:exp2d}, \ref{fig:exp3d}), providing reliable confidence quantification essential for decision-making in design contexts.
\end{itemize}
These demonstrated capabilities position GenPANIS as particularly well-suited for inverse design applications in multiphase materials. The ability to handle sparse, noisy observations combined with robust out-of-distribution generalization directly addresses key requirements for topology optimization \cite{bendsoe2003topology, sigmund2013topology}, multi-objective design \cite{deb2014multi,franck_multimodal_2017}, and microstructure engineering \cite{cecen2018versatile}. GenPANIS enables efficient design space exploration by: (i) sampling diverse candidates from the learned prior, (ii) rapid performance evaluation via the physics-aware surrogate, (iii) gradient-based refinement in latent space, and (iv) uncertainty quantification for decision support. The framework's robustness to configurations far from training data—a failure mode for purely data-driven surrogates \cite{lu2022comprehensive}—makes it suitable for automated materials discovery workflows \cite{hart2021machine, chen2021graph} where physical consistency must be guaranteed throughout iterative optimization.

Future work will extend the framework to incorporate design objectives and constraints explicitly, enabling direct multi-objective optimization in latent space. Additional directions include: scaling to three-dimensional problems where computational savings become even more critical and integration with experimental characterization for data-driven materials design \cite{kalidindi2015materials, dimiduk2018perspectives,zang2025psp}. The demonstrated ability to train effectively with minimal labeled data, combined with physics residuals, suggests particular promise for experimental settings where labeled data acquisition is expensive but physical principles are well-understood.

\bibliographystyle{unsrt}
\bibliography{extras/bibliography}

\appendix

\section{Definition of Signal to Noise Ratio (SNR)}
\label{appendix:snr}
For the generation of noisy observations in the inverse problem experiments, we assume additive Gaussian noise on the reference solution $\bs{u}_{ref}$ with observation model:
\begin{equation}
    \hat{\bs{u}} = \bs{u}_{ref} + \sigma_n \bs{\eta}, \quad \bs{\eta} \sim \mathcal{N}(\bs{0}, \bs{I}_{d_{obs}}),
\end{equation}
where $d_{obs}$ is the total number of observation points and $\sigma_n$ is the noise standard deviation. We specify the noise level in terms of the Signal-to-Noise Ratio (SNR) defined as:
\begin{equation}
    \text{SNR} = \frac{\|\bs{u}_{ref}\|_2}{\sigma_n \sqrt{d_{obs}}},
\end{equation}
which relates to the decibel scale via:
\begin{equation}
    \text{SNR}_{dB} = 20 \log_{10}(\text{SNR}) = 10 \log_{10}\left(\frac{\|\bs{u}_{ref}\|_2^2}{\sigma_n^2 d_{obs}}\right).
\end{equation}

\section{ Neural Network Architectures}
\label{appendix:shallowCNN}

In the current appendix, we present details about all the neural network architectures mentioned in the main sections.

\subsection{Normalizing Flow Prior}

Each affine coupling layer is defined using a fixed binary mask $\mathbf{m}_k \in \{0,1\}^L$, which partitions the input into transformed and non-transformed components. Given an input $\bz_{k-1}$, we define the masked input
\begin{equation}
\bz_{k-1}^{(m)} = \mathbf{m}_k \odot \bz_{k-1},
\end{equation}
where $\odot$ denotes elementwise multiplication. The superscript $(m)$ indicates that only the components selected by the mask $\mathbf{m}_k$ are retained for the subsequent scale and translation networks, while the remaining components remain untransformed. The forward transformation is then

\begin{equation}
\bz_k
=
\bz_{k-1}^{(m)}
+
(1 - \mathbf{m}_k) \odot
\left(
\bz_{k-1} \odot \exp(\mathbf{s}_k) + \mathbf{t}_k
\right),
\end{equation}

\noindent with

\begin{equation}
\mathbf{s}_k
=
\tanh\!\left(s_k(\bz_{k-1}^{(m)})\right)
\odot
\boldsymbol{\alpha}_k,
\qquad
\mathbf{t}_k = t_k(\bz_{k-1}^{(m)}),
\end{equation}
where $s_k(\cdot)$ and $t_k(\cdot)$ are neural networks, and $\boldsymbol{\alpha}_k \in \mathbb{R}^{d_z}$ is a learned per-dimension scaling parameter. The inverse mapping required for likelihood evaluation is 
\begin{equation}
\bz_{k-1}
=
\bz_k^{(m)}
+
(1 - \mathbf{m}_k) \odot
(\bz_k - \mathbf{t}_k) \odot \exp(-\mathbf{s}_k),
\end{equation}
and, due to the triangular structure of the Jacobian, the log-determinant of the inverse Jacobian is
\begin{equation}
\log
\left|
\det
\frac{\partial f_k^{-1}}{\partial \bz_k}
\right|
=
-
\sum_{i=1}^{L}
(1 - m_{k,i})\, s_{k,i}.
\end{equation}

Finally, the set of all trainable parameters induced by the flow is
\begin{equation}
\bt_p
=
\left\{
\bt_k^{(s)},\;
\bt_k^{(t)},\;
\boldsymbol{\alpha}_k
\right\}_{k=1}^{K},
\end{equation}

\noindent where $\bt_k^{(s)}$ and $\bt_k^{(t)}$ represent the weights and biases of the scale and translation networks, respectively. The binary masks $\mathbf{m}_k$ and the parameters of the base distribution are fixed and not learned. 

\begin{table}[H]
\centering
\begin{tabular}{|l|l|l|}
\hline
\textbf{Layers} & \textbf{Input $\rightarrow$ Output} & \textbf{Parameters} \\ \hline

Linear & 60 $\rightarrow$ 128 & 7808 \\ \hline
LeakyReLU & --- & --- \\ \hline
Linear & 128 $\rightarrow$ 128 & 16512 \\ \hline
LeakyReLU & --- & --- \\ \hline
Linear & 128 $\rightarrow$ 60 & 7740 \\ \hline

\textbf{Total} & --- & \textbf{32060} \\ \hline
\end{tabular}
\caption{Layers and parameters of the scale network $s_k(\cdot)$.}
\end{table}

\begin{table}[H]
\centering
\begin{tabular}{|l|l|l|}
\hline
\textbf{Layers} & \textbf{Input $\rightarrow$ Output} & \textbf{Parameters} \\ \hline

Linear & 60 $\rightarrow$ 128 & 7808 \\ \hline
LeakyReLU & --- & --- \\ \hline
Linear & 128 $\rightarrow$ 128 & 16512 \\ \hline
LeakyReLU & --- & --- \\ \hline
Linear & 128 $\rightarrow$ 60 & 7740 \\ \hline

\textbf{Total} & --- & \textbf{32060} \\ \hline
\end{tabular}
\caption{Layers and parameters of the translation network $t_k(\cdot)$.}
\end{table}

\begin{table}[H]
\centering
\begin{tabular}{|l|l|}
\hline
\textbf{Component} & \textbf{Parameters} \\ \hline

Scale network $s_k(\cdot)$ & 32060 \\ \hline
Translation network $t_k(\cdot)$ & 32060 \\ \hline
Learned scale parameter $\boldsymbol{\alpha}_k \in \mathbb{R}^{60}$ & 60 \\ \hline
Total per coupling layer & 64180 \\ \hline

\textbf{Total flow model parameters} & \textbf{770160} \\ \hline
\end{tabular}
\caption{Total parameters of the 12-coupling-layer RealNVP prior.}
\end{table}

\subsection{Effective Properties Network $f_{\bt_f}(\bz)$}

The details for the network $f_{\bt_f}(\bz)$ are shown in Table \ref{tab:NNdecoder}. The only type of activation function used is the Softplus activation function. These are used after each convolution/deconvolution layer and before the batch normalization layers, except for the last deconvolution layer, after which no activation function is applied.

\begin{table}[H]
\centering
\begin{tabular}{|l|l|l|l|}
\hline
\textbf{Layers} & \textbf{Feature Maps} & \textbf{Height $\times$ Width} & \textbf{Parameters} \\ \hline

Fully Connected Layer (60 $\rightarrow$ 16641) & --- & 16641 $\times$ 1 & 1015101 \\ \hline
Reshape & --- & 129 $\times$ 129 & --- \\ \hline

Convolution Layer (k3s2p1) & 16 & 65 $\times$ 65 & 160 \\ \hline
Batch Normalization & --- & 65 $\times$ 65 & 32 \\ \hline
Average Pooling Layer (k2s2) & --- & 32 $\times$ 32 & --- \\ \hline

Convolution Layer (k3s1p1) & 48 & 32 $\times$ 32 & 6960 \\ \hline
Batch Normalization & --- & 32 $\times$ 32 & 96 \\ \hline
Average Pooling Layer (k2s2) & --- & 16 $\times$ 16 & --- \\ \hline

Deconvolution Layer (k4s1p1) & 16 & 17 $\times$ 17 & 12304 \\ \hline
Batch Normalization & --- & 17 $\times$ 17 & 32 \\ \hline
Deconvolution Layer (k3s1p1) & --- & 17 $\times$ 17 & 145 \\ \hline

\textbf{Total} & 80 & --- & \textbf{1034830} \\ \hline
\end{tabular}
\caption{Layers and parameters of the neural network $f_{\bt_f}(\bz)$.}
\label{tab:NNdecoder}
\end{table}

\subsection{Encoder Network $g_{\bksi_g}(\hat{\bx}^{(i)})$}

The details for the network $g_{\bksi_g}(\hat{\bx}^{(i)})$ are shown in Table \ref{tab:NNencoder}. The only type of activation function used is the Softplus activation function. These are used after each convolution/deconvolution layer and before the batch normalization layers.

\begin{table}[H]
\centering

\begin{tabular}{|l|l|l|l|}
\hline
\textbf{Layers}         & \textbf{Feature Maps}  & \textbf{Height $\times$ Width} & \textbf{Parameters} \\ \hline
Input                       & ---           & 129 $\times$ 129                             &         ---                  \\ \hline
Convolution Layer (k3s2p1)       & 8          & 65 $\times$ 65                             &          80                  \\ \hline
Batch Normalization       & ---          & 65 $\times$ 65                             &            16                \\ \hline
Average Pooling Layer (k2s2)   & ---          & 32 $\times$ 32                             &            ---                \\ \hline
Convolution Layer (k3s1p1)       & 24          & 32 $\times$ 32                             &           1752                 \\ \hline
Batch Normalization       & ---          & 32 $\times$ 32                             &            48                \\ \hline
Average Pooling Layer (k2s2)   & ---          & 16 $\times$ 16                             &               ---             \\ \hline
Deconvolution Layer (k4s1p1)     & 8           & 17 $\times$ 17                             &         3080                   \\ \hline
Batch Normalization       & ---          & 17 $\times$ 17                             &            16                \\ \hline
Deconvolution Layer (k3s1p1)     & ---           & 17 $\times$ 17                             &          73                  \\ \hline
Flatten ($17 \times 17$ $\rightarrow$ 289)     & ---           & 289 $\times$ 1                             &          ---                  \\ \hline
Fully Connected Layer (289 $\rightarrow$ 60)    & ---           & 60 $\times$ 1                             &          17400                  \\ \hline
In total     & 40           & ---                             & 22465                        \\ \hline
\end{tabular}
\caption{Layers and parameters of the neural network $g_{\bksi_g}(\hat{\bx}^{(i)})$.}
\label{tab:NNencoder}
\end{table}

\section{Basis and Weight Functions for Virtual Observables}
\label{appendix:trialSolutions}

The PDE solution $u$ is represented using $d_{\by}=1024$ radial basis functions:
\begin{equation}
    \eta_i(\bs{s}) = \exp\left( -\frac{\lVert \bs{s} - \bs{s}_{i, 0} \rVert^2_2}{\Delta l^2}\right), \quad i = 1, \ldots, d_{\by},
    \label{eq:basisFuny}
\end{equation}
where the centers $\bs{s}_{i, 0}$ are located on a regular $32\times32$ grid over $\Omega$. Boundary conditions are enforced through the embedded coarse-grained solver (Section~\ref{sec:ydecoder}).

For virtual observables, we employ $K=289$ weight functions on a regular $17\times17$ grid:
\begin{equation}
    w_j(\bs{s}) = \tau(\bs{s}) \eta_j(\bs{s}), \quad \tau(\bs{s}) = s_1(1-s_1)s_2(1-s_2),
    \label{eq:basisFunw}
\end{equation}
where $\tau$ enforces homogeneous Dirichlet boundary conditions. Weighted residuals are evaluated using the trapezoidal rule on a $129\times129$ integration grid.







\end{document}